\documentclass[fleqn,10pt]{wlscirep}
\usepackage[utf8]{inputenc}
\usepackage[T1]{fontenc}
\usepackage{lineno}
\usepackage{algorithm, algorithmic}
\usepackage{multirow}
\usepackage[colorlinks=true]{hyperref}
\usepackage{ulem}  
\usepackage{pifont}
\usepackage{caption}  
\usepackage{subcaption}
\usepackage{forest}
\usepackage{textcomp}
\usepackage{xcolor}
\definecolor{darkGreen}{RGB}{22,173,2}
\definecolor{greenBlue}{RGB}{0,112,192}
\definecolor{shallowBlue}{RGB}{0,176,240}
\definecolor{darkPurple}{RGB}{216,77,206}
\newcommand{\darkGreen}[1]{{\color{darkGreen} #1}}
\newcommand{\greenBlue}[1]{{\color{greenBlue} #1}}
\newcommand{\shallowBlue}[1]{{\color{shallowBlue} #1}}

\newcommand{\red}[1]{{\color{red} #1}}
\newcommand{\blue}[1]{{{\color{blue} #1}}}

\newcommand{\orange}[1]{{{\color{orange} #1}}}  

\usepackage{CJKutf8}

\usepackage{notoccite}

\makeatletter

\renewenvironment{table}{
        \begingroup
        \@fileswfalse
        \begin{oldtable}
     }
     {
        \end{oldtable}
        \endgroup
     }
\makeatother

\makeatletter
\let\oldfigure\figure
\let\endoldfigure\endfigure
\renewenvironment{figure}{
        \begingroup
        \@fileswfalse
        \let\float@endH@ORI\float@endH
        \def\float@endH{after\float@endH@ORI}
        \oldfigure
     }
     {
        \endoldfigure
        \endgroup
     }
\makeatother

\makeatletter
\renewcommand\paragraph
  {%
    \@startsection{paragraph}{4}{\z@}{1.00ex \@plus 1ex \@minus .2ex}{-1ex}
      {\normalfont\normalsize\bfseries}
  }
\makeatother

\title{A large-scale dataset for end-to-end table recognition in the wild}

\author[1]{Fan Yang}
\author[1]{Lei Hu}
\author[2]{Xinwu Liu}
\author[1, 3,*]{Shuangping Huang}
\author[4]{Zhenghui Gu}
\affil[1]{School of Electronic and Information Engineering, South China University of Technology, Guangzhou, 510641, China}
\affil[2]{Zhuzhou CRRC Times Electric Co., Ltd, Zhuzhou, 412001, China}
\affil[3]{Pazhou Lab, Guangzhou, 510335, China}
\affil[4]{College of Automation Science and Engineering, South China University of Technology, Guangzhou, 510641, China}
\affil[*]{corresponding author(s): Shuangping Huang (eehsp@scut.edu.cn)}

\begin{abstract}
Table recognition (TR) is one of the research hotspots in pattern recognition, which aims to extract information from tables in an image. Common table recognition tasks include table detection (TD), table structure recognition (TSR) and table content recognition (TCR). TD is to locate tables in the image, TCR recognizes text content, and TSR recognizes spatial \& ontology (logical) structure. Currently, the end-to-end TR in real scenarios, accomplishing the three sub-tasks simultaneously, is yet an unexplored research area. One major factor that inhibits researchers is the lack of a benchmark dataset. To this end, we propose a new large-scale dataset named \uline{Tab}le \uline{Rec}ognition \uline{Set} (\textit{TabRecSet}) with diverse table forms sourcing from multiple scenarios in the wild, providing complete annotation dedicated to end-to-end TR research. It is the largest and first bi-lingual dataset for end-to-end TR, with 38.1K tables in which 20.4K are in English\, and 17.7K are in Chinese. The samples have diverse forms, such as the border-complete and -incomplete table, regular and irregular table (rotated, distorted, etc.). The scenarios are multiple in the wild, varying from scanned to camera-taken images, documents to Excel tables, educational test papers to financial invoices. The annotations are complete, consisting of the table body spatial annotation, cell spatial \& logical annotation and text content for TD, TSR and TCR, respectively. The spatial annotation utilizes the polygon instead of the bounding box or quadrilateral adopted by most datasets. The polygon spatial annotation is more suitable for irregular tables that are common in wild scenarios. Additionally, we propose a visualized and interactive annotation tool named \textit{TableMe} to improve the efficiency and quality of table annotation.
\end{abstract}

\begin{document}
\begin{CJK*}{UTF8}{gbsn}
\flushbottom
\maketitle

\thispagestyle{empty}


\section*{Background \& Summary}
Tables are commonly presented in images to organize and present information. To efficiently utilize information from table images, computer vision based pattern recognition techniques are used in table recognition (TR). It consists of three main tasks, table detection (TD), table structure recognition (TSR) and table content recognition (TCR), in relation to the localization of tables, the recognition of their internal structures, and the extraction of their text contents correspondingly.

\begin{figure}[htb!]
    \centering
    \includegraphics[width=1\linewidth, height=0.82\textwidth, keepaspectratio=false]{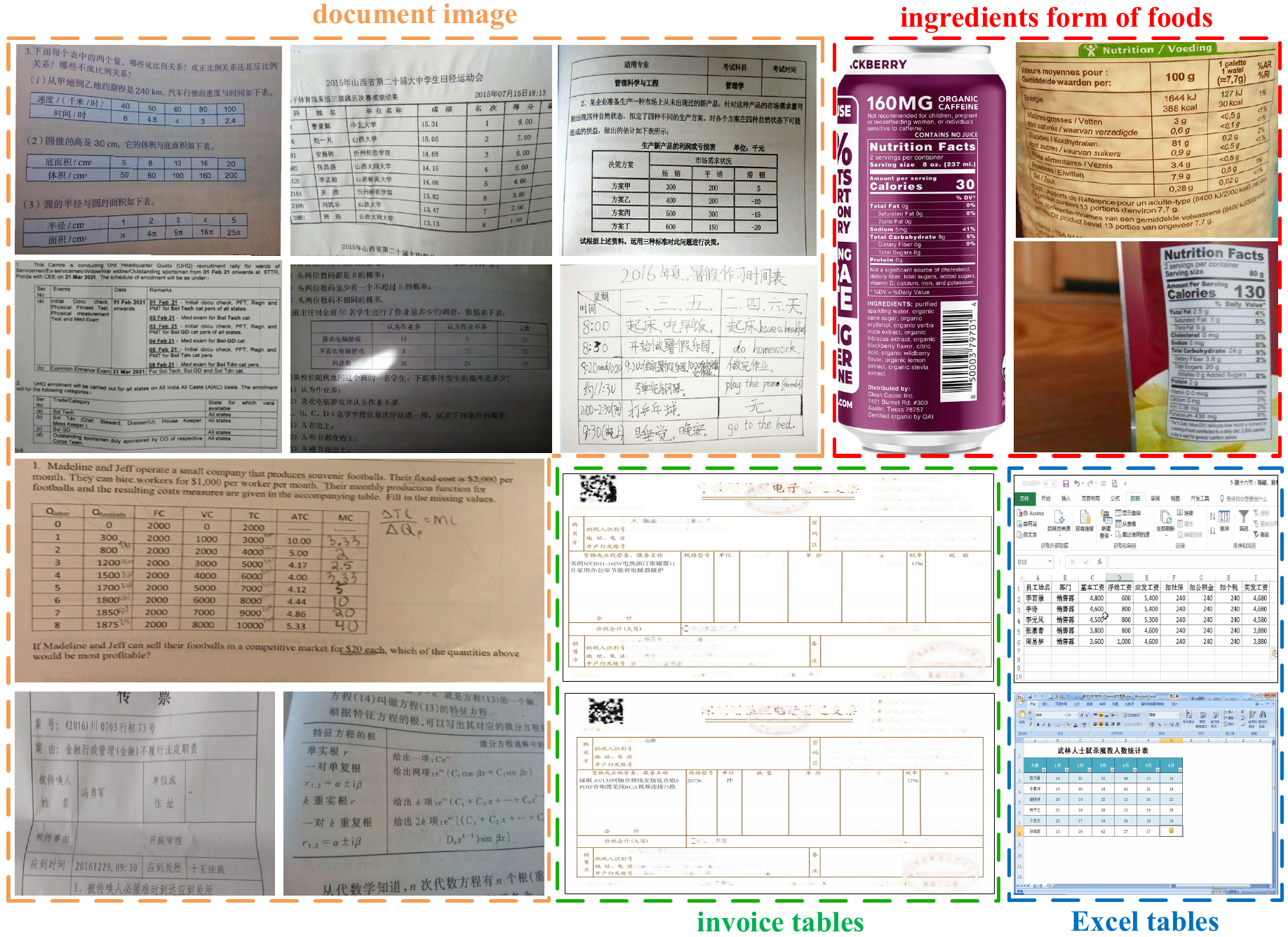}
    \caption{Some representative samples in \textit{TabRecSet}. The scenarios include the \orange{document images}, \red{ingredients form of foods}, \greenBlue{Excel tables} and \darkGreen{invoice tables}. Because of the page distortions or camera views, most tables are irregular, i.e., with rotations, inclinations, concave/convex/wrinkle distortions, etc. Some special table forms are exhibited, e.g., the nested table, under- and over-exposed table, border-incomplete table, table with hand-written contents and hand-drawn table.}
    \label{fig: dataset overview}
\end{figure}

Currently, the end-to-end TR task in real scenarios, with the purpose of fulfilling all three sub-tasks simultaneously, is yet unexplored. One major factor that inhibits researchers is the lack of a well-rounded benchmark dataset. For instance, as shown in Table~\ref{tab: dataset statistical summary and comparison}, early (before 2018) TR datasets, such as UNLV~\cite{DBLP:conf/das/ShahabSKD10}, ICDAR13~\cite{DBLP:conf/icdar/2013} and ICDAR17~\cite{DBLP:conf/icdar/GaoYJHT17}, only contain a few samples (less than 2.5k). Later, large-scale TR datasets~\cite{DBLP:conf/jcdl/SiegelLPA18,DBLP:conf/icdar/ZhongTJ19,DBLP:journals/corr/abs-1908-04729,DBLP:conf/icdar/DengRM19,DBLP:conf/lrec/LiCHWZL20,DBLP:conf/eccv/ZhongSJ20,10.1007/978-3-030-86331-9_36,smock2022pubtables}~were proposed since 2019, but the annotations are generated by programs instead of human involved and only scanned regular tables are included, hindering the diversity of the datasets due to the monotonous backgrounds and spatial features (e.g. without rotation, distortion, etc.). In fact, to enrich the diversity, it is necessary to collect data in various real scenarios. For example, Gao \normalem\emph{et al.}~\cite{DBLP:conf/icdar/GaoHDMYFKL19}~proposed a dataset named ICDAR19 for TD and TSR tasks. It is the first real dataset in the historical document scenario, yet its volume is small (2.4k images). Until recently, Long \normalem\emph{et al.}~\cite{9710258}~proposed a large-scale (14.5k) practical dataset WTW that covers multiple scenarios in the wild. Although WTW is the largest and multi-scenario, it is only suitable for the TSR task as it lacks table location and content annotations. Furthermore, quadrilateral box is used in the cell location annotation, which is imprecise to distorted (caused by folds and bends in a paper) table images. Overall, we conclude three main drawbacks of these datasets as follows: 1. Only provide annotations for sub-tasks (TD, TSR and TCR), which are not complete for the end-to-end TR task. 2. Either the scales are small, or scenario diversities are limited. 3. Only have the Bounding box (Bbox) or quadrilateral as the spatial annotation that cannot flexibly adapt to the shape changes. 

\begin{table}[htb]
\begin{center}
\setlength{\tabcolsep}{0.85mm}
\renewcommand\arraystretch{1.4}
\begin{tabular}{|c|c|c|cccc|c|c|c|c|c|}
\hline
\multirow{2}{*}{\textbf{Dataset}}                          & \multirow{2}{*}{\textbf{\#Images}} & \multirow{2}{*}{\textbf{\#Tables}} & \multicolumn{4}{c|}{\textbf{Task}}                                                                                                                                          & \multirow{2}{*}{\textbf{\begin{tabular}[c]{@{}c@{}}Multiple\\Wild\\Scenarios\end{tabular}}} & \multirow{2}{*}{\textbf{\begin{tabular}[c]{@{}c@{}}Spatial\\Annotation\\Flexibility\end{tabular}}} & \multirow{2}{*}{\textbf{\begin{tabular}[c]{@{}c@{}}Border-\\incomplete\\Diversity\end{tabular}}} & \multirow{2}{*}{\textbf{Bi-lingual}} & \multirow{2}{*}{\textbf{Year}} \\ \cline{4-7}
                                                           &                                    &                                    & \multicolumn{1}{c|}{\textbf{TD}} & \multicolumn{1}{c|}{\textbf{TSR}} & \multicolumn{1}{c|}{\textbf{TCR}} & \textbf{\begin{tabular}[c]{@{}c@{}}End-to-End\\ TR\end{tabular}} &                                                                                                       &                                                                                                      &                                                                                              &                                      &                                \\ \hline
UNLV~\cite{DBLP:conf/das/ShahabSKD10}                                                       & 427                                & 558                                & \multicolumn{1}{c|}{\ding{52}}           & \multicolumn{1}{c|}{\ding{52}}            & \multicolumn{1}{c|}{\ding{52}}            & \ding{56}                                                                & \ding{56}                                                                                                     & \ding{56}                                                                                                    & \ding{56}                                                                                            & \ding{56}                                    & 2010                           \\ \hline
ICDAR13~\cite{DBLP:conf/icdar/2013}                                                    & 128                                & 156                                & \multicolumn{1}{c|}{\ding{52}}           & \multicolumn{1}{c|}{\ding{52}}            & \multicolumn{1}{c|}{\ding{52}}            & \ding{56}                                                                & \ding{56}                                                                                                     & \ding{56}                                                                                                    & \ding{56}                                                                                            & \ding{56}                                    & 2013                           \\ \hline
ICDAR17~\cite{DBLP:conf/icdar/GaoYJHT17}                                                    & 2417                               & 1020                               & \multicolumn{1}{c|}{\ding{52}}           & \multicolumn{1}{c|}{\ding{56}}            & \multicolumn{1}{c|}{\ding{56}}            & \ding{56}                                                                & \ding{56}                                                                                                     & \ding{56}                                                                                                    & \ding{56}                                                                                            & \ding{56}                                    & 2017                           \\ \hline
DeepFigures~\cite{DBLP:conf/jcdl/SiegelLPA18}                                                & 1.67M                              & 1.4M                               & \multicolumn{1}{c|}{\ding{52}}           & \multicolumn{1}{c|}{\ding{56}}            & \multicolumn{1}{c|}{\ding{56}}            & \ding{56}                                                                & \ding{56}                                                                                                     & \ding{56}                                                                                                    & \ding{56}                                                                                            & \ding{56}                                    & 2018                           \\ \hline
PubLayNet~\cite{DBLP:conf/icdar/ZhongTJ19}                                                  & 362K                               & 113K                               & \multicolumn{1}{c|}{\ding{52}}           & \multicolumn{1}{c|}{\ding{56}}            & \multicolumn{1}{c|}{\ding{56}}            & \ding{56}                                                                & \ding{56}                                                                                                     & \ding{56}                                                                                                    & \ding{56}                                                                                            & \ding{56}                                    & 2019                           \\ \hline
SciTSR~\cite{DBLP:journals/corr/abs-1908-04729}                                                     & 15K                                & 15K                                & \multicolumn{1}{c|}{\ding{56}}           & \multicolumn{1}{c|}{\ding{52}}            & \multicolumn{1}{c|}{\ding{52}}            & \ding{56}                                                                & \ding{56}                                                                                                     & \ding{56}                                                                                                    & \ding{56}                                                                                            & \ding{56}                                    & 2019                           \\ \hline
Table2Latex~\cite{DBLP:conf/icdar/DengRM19}                                                & 465K                               & 465K                               & \multicolumn{1}{c|}{\ding{56}}           & \multicolumn{1}{c|}{\ding{52}}            & \multicolumn{1}{c|}{\ding{52}}            & \ding{56}                                                                & \ding{56}                                                                                                     & NA                                                                                                   & \ding{56}                                                                                            & \ding{56}                                    & 2019                           \\ \hline
ICDAR19~\cite{DBLP:conf/icdar/GaoHDMYFKL19}                                                    & 2,439                              & 3.6K                               & \multicolumn{1}{c|}{\ding{52}}           & \multicolumn{1}{c|}{\ding{52}}            & \multicolumn{1}{c|}{\ding{56}}            & \ding{56}                                                                & \ding{56}                                                                                                     & \ding{56}                                                                                                    & \ding{56}                                                                                            & \ding{56}                                    & 2019                           \\ \hline
\multirow{2}{*}{TableBank~\cite{DBLP:conf/lrec/LiCHWZL20}}                                 & 278K                               & 417K                               & \multicolumn{1}{c|}{\ding{52}}           & \multicolumn{1}{c|}{\ding{56}}            & \multicolumn{1}{c|}{\ding{56}}            & \ding{56}                                                                & \ding{56}                                                                                                     & \ding{56}                                                                                                    & \ding{56}                                                                                            & \ding{56}                                    & \multirow{2}{*}{2020}          \\ \cline{2-11}
                                                           & 145K                               & 145K                               & \multicolumn{1}{c|}{\ding{56}}           & \multicolumn{1}{c|}{\ding{52}}            & \multicolumn{1}{c|}{\ding{56}}            & \ding{56}                                                                & \ding{56}                                                                                                     & NA                                                                                                   & \ding{56}                                                                                            & \ding{56}                                    &                                \\ \hline
PubTabNet~\cite{DBLP:conf/eccv/ZhongSJ20}                                                  & 568K                               & 568K                               & \multicolumn{1}{c|}{\ding{56}}           & \multicolumn{1}{c|}{\ding{52}}            & \multicolumn{1}{c|}{\ding{52}}            & \ding{56}                                                                & \ding{56}                                                                                                     & \ding{56}                                                                                                    & \ding{56}                                                                                            & \ding{56}                                    & 2020                           \\ \hline
TableX~\cite{10.1007/978-3-030-86331-9_36}                                                     & 1M+                                & 1M+                                & \multicolumn{1}{c|}{\ding{56}}           & \multicolumn{1}{c|}{\ding{52}}            & \multicolumn{1}{c|}{\ding{52}}            & \ding{56}                                                                & \ding{56}                                                                                                     & NA                                                                                                   & \ding{56}                                                                                            & \ding{56}                                    & 2021                           \\ \hline
PubTables-1M~\cite{smock2022pubtables}                                               & 1M+                                & 1M+                                & \multicolumn{1}{c|}{\ding{52}}           & \multicolumn{1}{c|}{\ding{52}}            & \multicolumn{1}{c|}{\ding{56}}            & \ding{56}                                                                & \ding{56}                                                                                                     & \ding{56}                                                                                                    & \ding{56}                                                                                            & \ding{56}                                    & 2021                           \\ \hline
WTW~\cite{9710258}                                                        & 14.5K                              & 14.5K                              & \multicolumn{1}{c|}{\ding{56}}           & \multicolumn{1}{c|}{\ding{52}}            & \multicolumn{1}{c|}{\ding{56}}            & \ding{56}                                                                & \ding{52}                                                                                                     & \ding{56}                                                                                                    & \ding{56}                                                                                            & \ding{52}                                    & 2021                           \\ \hline
\begin{tabular}[c]{@{}c@{}}TabRecSet\\ (Ours)\end{tabular} & 32.07K                             & 38.17K                             & \multicolumn{1}{c|}{\ding{52}}           & \multicolumn{1}{c|}{\ding{52}}            & \multicolumn{1}{c|}{\ding{52}}            & \ding{52}                                                                & \ding{52}                                                                                                     & \ding{52}                                                                                                    & \ding{52}                                                                                            & \ding{52}                                    &                                \\ \hline
\end{tabular}
\end{center}
\caption{A statistical summary and comparison between our \textit{TabRecSet} dataset and the existing datasets.\textbf{End-to-End TR}: Extract the table body position, structure, and content simultaneously from a complete image. \textbf{Spatial annotation flexibility}: The dataset uses the polygon instead of the Bounding box (Bbox) or quadrilateral to annotate the table or cell position. \textbf{Border-incomplete Diversity}: The dataset has multiple types of border-incomplete tables. \textbf{NA}: This item for the dataset is not applicable.}
\label{tab: dataset statistical summary and comparison}
\end{table}

We propose a dataset named \uline{Tab}le \uline{Rec}ognition \uline{Set} (\textit{TabRecSet}) with samples exhibited in Figure~\ref{fig: dataset overview}. \textbf{\uline{To the best of our knowledge, it is the largest and most well-rounded real dataset, collecting data from various wild scenarios with diverse table styles and complete \& flexible annotation against for the end-to-end TR task with the purpose of filling the gap in this research area.}} \textbf{Large Scale:} The data volume (including more than 38,100 real table images) is 2.6 times larger than the largest known dataset WTW. \textbf{Wild Scenario:} Data are collected via scanners or cameras in various wild scenarios including documents, Excel tables, exam papers, financial invoices, etc. \textbf{Robust Diversity:} It contains different table forms, such as the regular and irregular table (rotated, distorted, etc.), border-complete (all-line) and -incomplete table. The distortions of irregular tables may severely break the spatial alignment of rows, columns and cells, increasing the difficulty of the TSR task. The recognition of border-incomplete tables such as the three-\cite{Tables_Requirements} and no-line (without borders) tables are also more difficult and challenging. \textbf{Completeness:} In order to provide a complete annotation for the {TR} task, the annotation of every table sample in \textit{TabRecSet} contains its body and cell location as well as its structure and content. \textbf{Flexibility:} In order to provide accurate and precise annotations to distorted tables, \textit{TabRecSet} uses polygons instead of bounding boxes to annotate the outside and inside table borders. \textbf{Bi-lingual:} \textit{TabRecSet} contains Chinese and English tables with a proportion of 46.5\% and 53.5\% independently.

In addition, since the process of dataset building is quite time-consuming, we developed a visualized and interactive annotation tool named \textit{TableMe} to speed up the annotation process and ensure data quality. We also designed several automatic techniques, such as the auto annotating of table structures and the automatic generation of three- and no-line tables, to benefit the end-to-end TR task in wild scenarios.

\section*{Methods}



\begin{figure}[htb]
    \centering
    \includegraphics[width=1\linewidth]{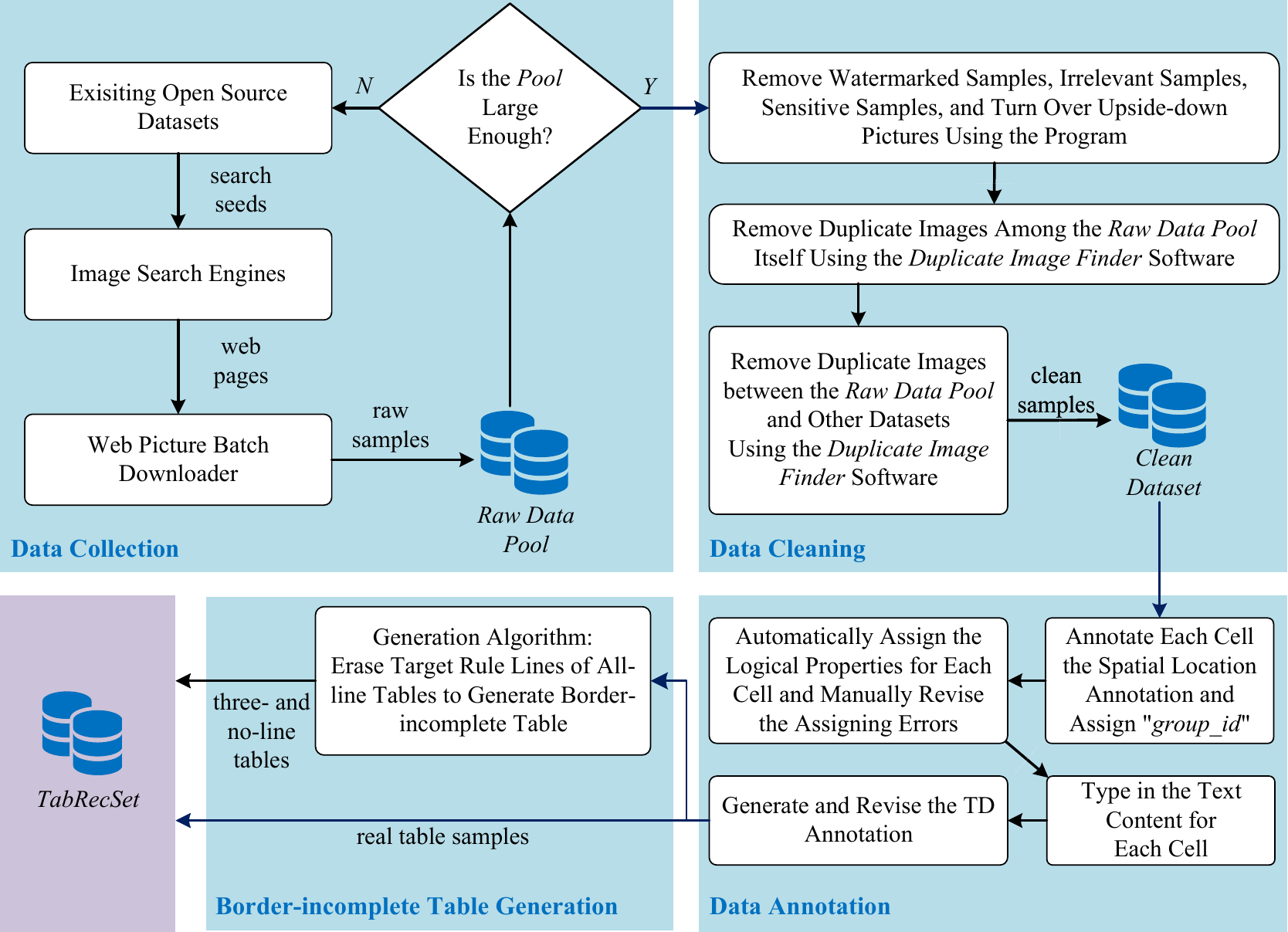}
    \caption{The creation flow chart of TabRecSet. The data collection aims to collect raw image samples and outputs a \textit{Raw Data Pool}, which stores candidate data samples. The data cleaning step generates clean samples from \textit{Raw Data Pool} and gathers them into a \textit{Clean Dataset}. In the data annotation step, we use \textit{TableMe} to annotate the clean sample and save the annotation in the \textit{TabRecSet} annotation format. This step is aided by several auto-annotation algorithms to improve efficiency. The border-incomplete table generation step aims to enlarge the scale \textit{TabRecSet} by our proposed three-line table generating algorithm.}
    \label{fig: Data creation flow chart}
\end{figure}

In this section, we elaborate on all the details in the \textit{TabRecSet} creation procedure, which ensures the quality, reproducibility, and creation efficiency of the dataset. As Figure~\ref{fig: Data creation flow chart}~illustrates, this procedure mainly consists of four steps. The first three steps, data collection, data cleaning, and data annotation, following a normal and standard procedure of building most datasets, output all border-complete table samples. Particularly, to increase the number of samples and the variety of table styles, they are also used to generate border-incomplete tables (e.g., three- or no-line tables) in the fourth step. The generation process is automatic and image-based, which basically replaces border pixels from border-complete table images with its background pixels.

\begin{figure}[htb]
    \begin{center}
       \includegraphics[width=1\linewidth]{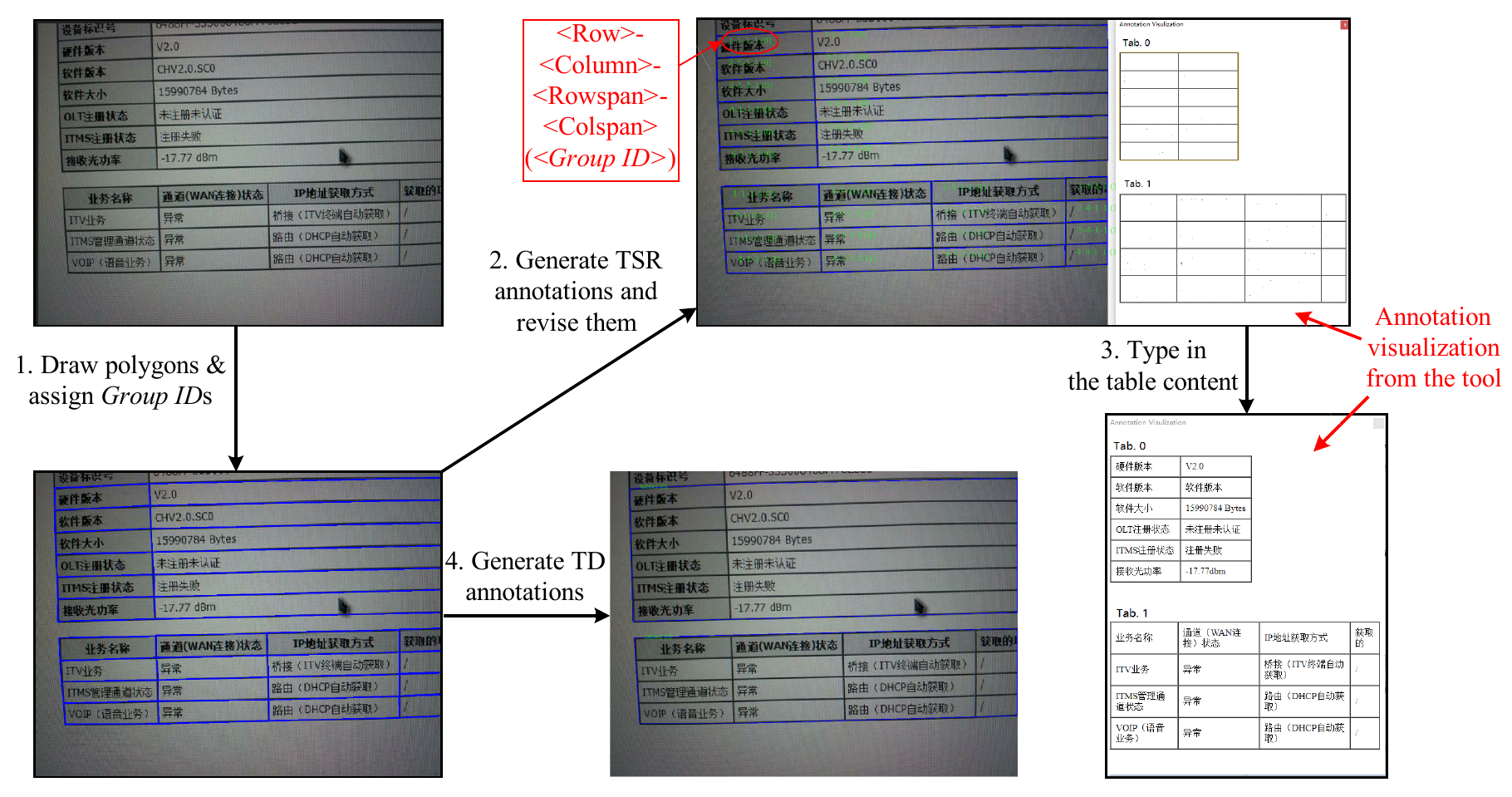}
        \caption{An intuitive illustration of the data annotation step showed in Figure~\ref{fig: Data creation flow chart}. Please zoom in for details.}
        \label{fig: Data annotating diagram}
    \end{center}
\end{figure}

\subsection*{Data Collection}
\label{subsec: data collection}
The general purpose of the data collection step is to build a raw data pool by searching and downloading enough table-related images through the Internet. Firstly, we randomly pick camera-taken table image samples from open source datasets such as WTW or \textit{Tal ocr\_table}~\cite{table_survey} as search seeds. Then they are input into search engines (e.g., Google or Baidu) with the Usage Rights filter enabled and return plenty of similar images that comply with the Creative Commons licenses. For the search engine that does not have the Usage Rights filter, we manually open the image's original source to check whether the image complies with the licenses. After that, the search results are downloaded via a web page-based image downloader called \textit{ImageAssistant}~\cite{imageAssistant}. The whole process stops when the total size of the downloaded images exceeds a specified threshold. The advantage of using search engines to collect table images is that the search result covers a wide variety of data collection scenarios, including reports, exam papers, documents, invoices, books, etc. In addition, to further increase data diversity, we use raw images rather than keywords as the search seeds because the image content does not play an important role in the way of search by image, and the search engine searches for matched images based on the pixel-level similarity of input images. In this way, the search engine will return a large number of table images with a diverse range of formats and styles, covering many special cases such as irregular, distorted, and incomplete tables, while in the way of search by keyword, the extent and degree of the distortion or the incompleteness are difficult to describe. As shown in Figure~\ref{fig: Data creation flow chart}, we repeat the search and download process until the data size of the \textit{Raw Data Pool} exceeds the expected dataset scale by a pre-specified margin based on the filtering rate in the cleaning step.

\subsection*{Data Cleaning}
The data cleaning step in Figure~\ref{fig: Data creation flow chart} is a human-involved process of fixing or removing incorrect, incomplete, incorrectly formatted, duplicate or irrelevant data within the \textit{Raw Data Pool}.

Since the raw data is collected through the Internet in the way of search by image, some images that do not include actual table instances may be returned by the search engine. These incorrect and table irrelevant data are removed in the first place. Meanwhile, watermarked images are removed because of copyright protection. In addition, for privacy considerations, we also remove sensitive information, such as location, ID, phone number, etc., from those images. In terms of the data format, in order to keep the variety of table format and styles, we only fix sideways or upside down images. After that, we detect and remove duplicate images (keep the image with the highest image resolution and remove the rest) via \textit{Duplicate Image Finder~\cite{Duplicate-Image-Finder}} software. It helps the user identify duplicate images, even if they are resized, edited, flipped, color-corrected, etc., by grouping them together. Moreover, samples duplicated from other datasets without a derivative license, e.g. the \textit{Tal ocr\_table} dataset, are also entirely removed.

The data cleaning step filters out approximately 30\% (based on our experience) of “dirty” data from \textit{Raw Data Pool} to obtain a clean dataset with high-quality data for the annotating process in the following step.

\subsection*{Data Annotation}
\label{subsec: Format, Tool and Data Annotation}
We first introduce the annotation format of \textit{TabRecSet} (\hyperref[subsubsec: data annotation format]{Annotation Format}) and our developed annotation tool \textit{TableMe} (\hyperref[subsubsec: Data Annotation Tool]{Annotation Tool}). Then, we propose a TSR auto-annotating algorithm (\hyperref[subsubsec: TSR Annotation Generating Algorithm]{TSR Auto-annotating Algorithm}) to automatically generate logical structure annotation based on the spatial structure annotation. Finally, in the \hyperref[subsubsec: data annotation step]{Annotation Step} subsection, we describe the data annotation step for the \textit{Clean Dataset}, which is mainly performed on our tool, including cell polygons drawing, algorithm-assisted logical locations generation, typing in text contents and table body polygons generation.

\subsubsection*{Annotation Format}
\label{subsubsec: data annotation format}
The complete annotation for the end-to-end TR task is complex as it includes table body position for the TD task, the cell spatial \& logical location for the TSR task and the cell text content for the TCR task, covering multiple heterogeneous information from spatial and logical to text data. It is necessary to utilize proper annotation formats to organize the information coherently and concisely. We choose the \textit{LabelMe}~\cite{russell2008labelme}~annotation format as the framework for our annotation formats since this framework supports compactly organizing the heterogeneous table information. This annotation format framework is defined by the fields and sub-fields listed in Table~\ref{tab: fields of LabelMe format} and Table~\ref{tab: sub-fields of the shapes field}.

\begin{table}[htb]
\begin{center}
\setlength{\tabcolsep}{0.95mm}
\renewcommand\arraystretch{0.8}
\begin{tabular}{|c|c|c|c|c|c|}
\hline
\multirow{2}{*}{\textbf{Field Name}} & \multirow{2}{*}{\textbf{Description}}                                          & \multirow{2}{*}{\textbf{Field Name}} & \multirow{2}{*}{\textbf{Description}}                                                    & \multirow{2}{*}{\textbf{Field Nme}} & \multirow{2}{*}{\textbf{Description}} \\
                                     &                                                                                &                                      &                                                                                          &                                     &                                       \\ \hline
version                              & the version of \textit{LabelMe}                                                         & lineColor                            & \begin{tabular}[c]{@{}c@{}}color of the lines\\ in annotation objects\end{tabular}       & imageData                           & encoded image data                    \\ \hline
flags                                & flags of the image                                                             & fillColor                            & \begin{tabular}[c]{@{}c@{}}color of the regions of\\ the annotation objects\end{tabular} & imageHeight                         & height of the image                   \\ \hline
shapes                               & \begin{tabular}[c]{@{}c@{}}the annotation objects\\ for the image\end{tabular} & imagePath                            & file path of the image                                                                   & imageWidth                          & width of the image                    \\ \hline
\end{tabular}
\end{center}
\caption{Fields of LabelMe format framework.}
\label{tab: fields of LabelMe format}
\end{table}

\begin{table}[htb]
\begin{center}
\setlength{\tabcolsep}{0.85mm}
\renewcommand\arraystretch{0.75}
\begin{tabular}{|c|c|c|c|c|c|}
\hline
\multirow{2}{*}{\textbf{Field Name}} & \multirow{2}{*}{label}                                                   & \multirow{2}{*}{points}                                                 & \multirow{2}{*}{group\_id}                                                     & \multirow{2}{*}{shape\_type}                                                                                & \multirow{2}{*}{flags}                                                   \\
                                     &                                                                          &                                                                         &                                                                                &                                                                                                             &                                                                          \\ \hline
\textbf{Description}                 & \begin{tabular}[c]{@{}c@{}}class of the\\ annotation object\end{tabular} & \begin{tabular}[c]{@{}c@{}}coordinates of the\\ annotation\end{tabular} & \begin{tabular}[c]{@{}c@{}}instance id of the\\ annotation obejct\end{tabular} & \begin{tabular}[c]{@{}c@{}}type of the \\ annotation object\\ (polygon, circle,\\ line, point)\end{tabular} & \begin{tabular}[c]{@{}c@{}}flags of the\\ annotation object\end{tabular} \\ \hline
\end{tabular}
\end{center}
\caption{The sub-fields of the \textit{"shapes"} field.}
\label{tab: sub-fields of the shapes field}
\end{table}

We form two annotation formats: a table-wise annotation format in which the annotation object is a table and a cell-wise one with a cell as the annotation object. The table-wise annotation format is for the TD task containing the spatial location information for the table. The cell-wise format contains the spatial \& logical locations and text content information for a cell annotation, and the collection of these cell annotations completely describes TSR and TCR annotations for the whole table~\cite{DBLP:journals/corr/abs-2106-10598}. 

\begin{figure}[htb]
    \begin{center}
       \includegraphics[width=0.9\linewidth]{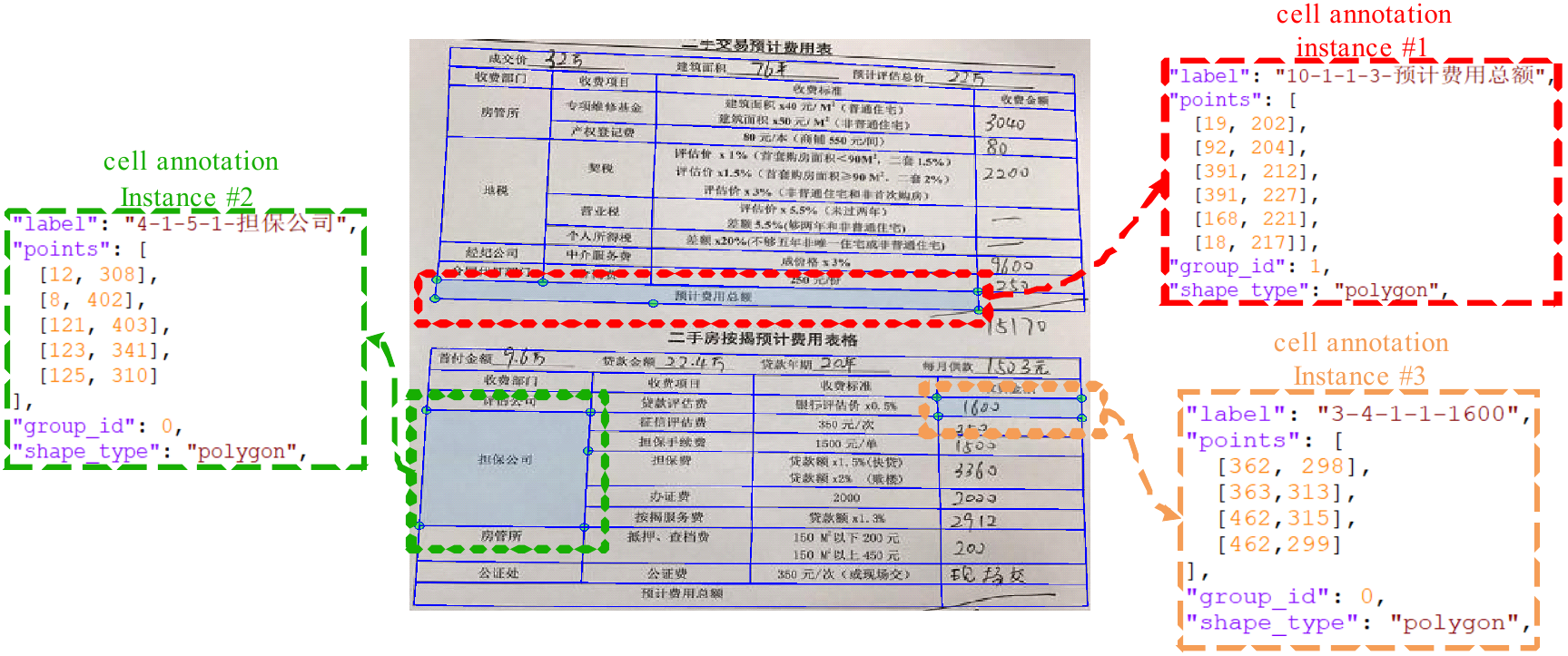}
       \caption{Three annotation instances in the cell-wise annotation format.}
       \label{fig: An annotation example in the TabRecSet annotation format}
    \end{center}
\end{figure}

Concretely, the table-wise annotation format utilizes the \textit{"points"} field to store the vertexes coordinates of the table body polygon and the same \textit{"group\_id"} (an integer) to distinguish different table instances. The cell-wise format uses the \textit{"label"} field to store a text string that encodes the cell's logical locations and text content, the \textit{"points"} field to store the vertexes coordinates of the polygon along the border of the cell, and the \textit{"group\_id"} field as in the table-wise format to mark the cell to which table instance it belongs. The text string in the \textit{"label"} field is in the form of \textit{"<Row>-<Column>-<Rowspan>-<Colspan>-<Text content>"} in which the <Row> (<Column>) means the row (column) number of the cell and the <Rowspan> (<Colspan>) means how many rows (columns) the cell spans. The row number, column number, rowspan and colspan are also called logical properties for short. Figure~\ref{fig: An annotation example in the TabRecSet annotation format} gives examples of three cell annotation instances in the cell-wise annotation format. The \red{\#1 instance} indicates the cell annotation object located by polygon $[[19,202], [92,204], [391,212],$ $[391,227], [168,221], [18,217]]$, in the tenth row and the first column of table 1 (group id=1), spanning three columns, with the text content "预计费用总额". The \darkGreen{\#2 instance} indicates the cell annotation object located by polygon $[[12,308], [8,402], [121,403],$ $[123,341], [125,310]]$, in the fourth row and the first column of the table 0, spanning five rows, with the text content "担保公司". The \orange{\#3 instance} indicates the cell annotation object located by polygon $[[362,298], [363,313], [462,315],$ $[462,299]]$, in the third row and the fourth column of the table 0, with the \textbf{\uline{hand-written "1600" as the text content}}.

\subsubsection*{Annotation Tool}
\label{subsubsec: Data Annotation Tool}
\textit{TableMe} originate from the famous annotation tool~\textit{LabelMe}, which is powerful in providing the annotation for the image segmentation task, and \textit{TableMe} completely inherits this feature leading to a great capacity for the table or cell's spatial annotating. Besides, it possesses annotating functions for the table structure \& content and supports assigning logical properties and \textit{"group\_id"} for a group of selected cell annotations, enhancing the structure annotating speed significantly. Most amazingly, it can intuitively visualize the logical structure and content of the table, which helps us to transcribe the text content to the annotation straightforwardly and efficiently (as shown in Fig~\ref{fig: Data annotating diagram}).

\begin{figure}[htb]
    \centering
    \includegraphics[width=1\linewidth]{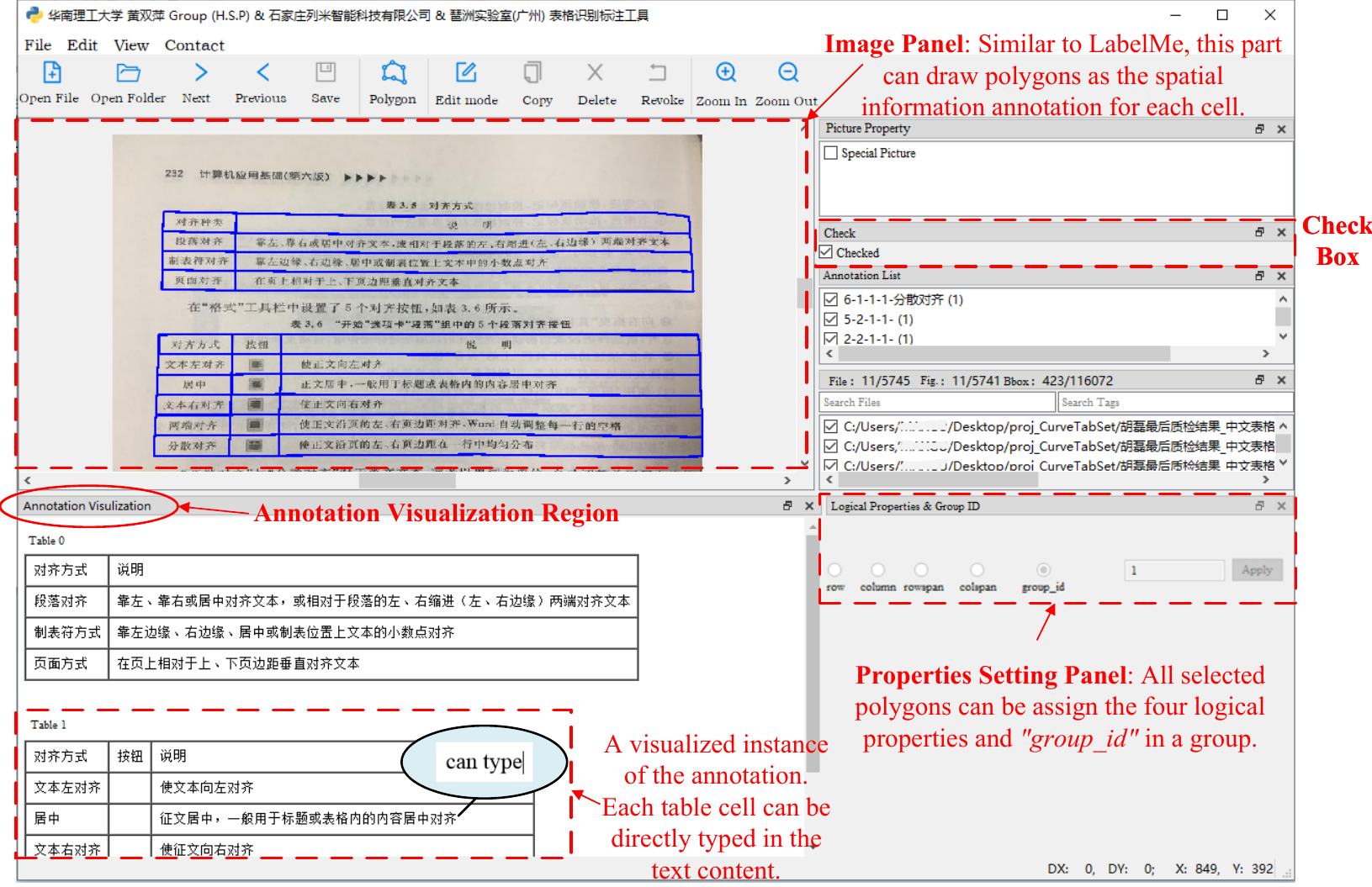}
    \caption{The main interface of \textit{TableMe}. Please zoom in for details.}
    \label{fig: annotating tool main interface}
\end{figure}

As illustrated in Figure~\ref{fig: annotating tool main interface}, \textit{TableMe} is mainly composed of three parts: an image panel (upper-left), a setting panel for polygon properties (structure logical \& \textit{"group\_id"})  (lower-right) and an annotation visualization region (lower-left). The image panel is a feature originated from the \textit{LabelMe}, which not only inherits the convenient polygons drawing functions of \textit{LabelMe} for the table/cell position annotating but also supports selecting these polygons in a group for the setting of polygon properties in the setting panel. 

The properties setting panel includes five options, i.e., the row, column, rowspan, colspan and \textit{"group\_id"}, and two widgets, i.e., an editable text for the option value input and an "Apply" button. For example, for the setting of the row property, the annotator can first select the polygons on the same row in the image panel, choose the "row" option, input the row number, and click the "Apply" button to confirm. The other properties can be set with totally the same operations. With this feature, \textit{TableMe} enables an intuitive annotating way for the table structure, which is high-efficiency and has less error tendency when people annotate.

The annotation visualization region supports two functions: visualizing the table structure and content annotation in a digital table form and annotating the table content directly in the digital table. After annotating the table/cell position in the image panel and table structure in the properties setting panel, this region will immediately show digitalized tables with the same logical structure as the tables in the image. Amazingly, as these digital tables are interactable, it allows users to annotate the text content of a cell simply by clicking that digital cell in this region and typing in the text string directly. This feature, on the one hand, increases the speed of annotating the text content and, on the other hand, provides an effective way for users to check whether the manually or automatically generated structure annotations are correct or not.

In the multiple tables case, the annotator can set the polygons in the same table with the same \textit{"group\_id"} and ensure polygons in different tables have different \textit{"group\_id"} by operating the widgets on the properties setting panel. When finished setting the \textit{"group\_id"} for each polygon, as shown in the figure, the annotation of multiple tables will be visualized in the annotation visualization region and distinguished from each other via table numbers that are equal to the \textit{"group\_id"}s. 

Besides above mentioned three parts, there is a check box named "Checked" on the right side of the interface. When an image is annotated, we can check this box to mark the image as annotated, and the value of the \textit{"flags"} field in~\ref{tab: fields of LabelMe format}~will be set to "true". The annotation is saved as a \textit{LabelMe} JSON file in the \textit{TabRecSet} annotation format with the same filename of the image but a different file extension. After finishing annotating all images in \textit{Clean Dataset}, a folder has paired JSON files and JPG files can be obtained, which contains the complete TR annotation for each image.

In conclusion, in terms of functionality, \textit{TableMe} is dedicated to annotating the end-to-end TR task in the wild scenario as it supports annotating table position, structure and content in multiple tables and irregular table cases. In terms of efficiency, it highly improves the speed of table annotating, annotation checking and revising, especially when the image contains many tables and the table has a large cell number, as it avoids trivially treating each table cell one by one.


 

\subsubsection*{TSR Auto-annotating Algorithm}
\label{subsubsec: TSR Annotation Generating Algorithm}
The TSR annotation generating algorithm consists of two parts: 1) A table image rectification process to eliminate the irregularity of tables (distortions, rotations, etc.) and 2) a logical property computing process to compute the logical properties for each cell. Without the table irregularities, the logical property computing process can get the most out of it, thus making this algorithm has a high computing accuracy, essentially preventing us from manually annotating these properties and improving efficiency.

\begin{figure}[htb]
    \begin{center}
       \includegraphics[width=1\linewidth]{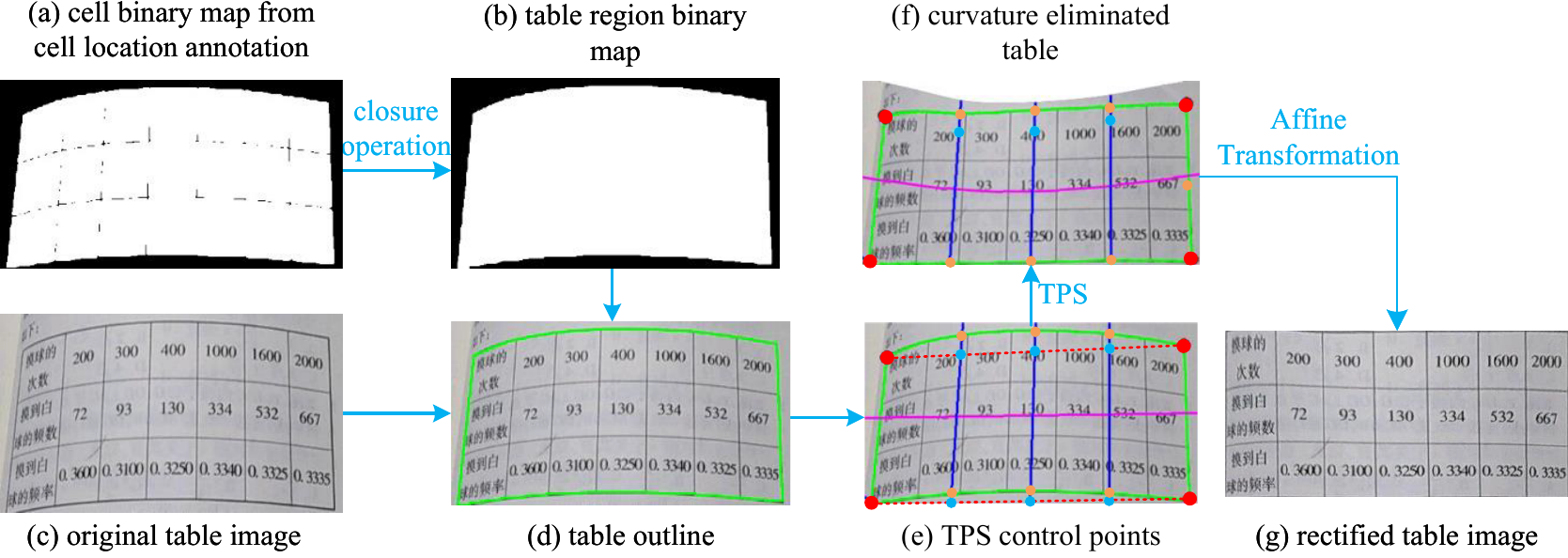}
        \caption{The table image rectification algorithm. The \shallowBlue{blue}, \red{red}, \orange{orange} points, \blue{blue lines} and two \red{red dashed lines} in Fig. (e) (f) are the TPS target points, TPS source points, corner points of the table outline, normal lines and corner lines, respectively. A {\color{darkPurple} purple line} is drawn horizontally in the middle of Fig. (e) and is distorted in Fig. (f), visualizing the extent and direction of the TPS transformation.}
        \label{fig: table image rectification process}
    \end{center}
\end{figure}

The rectification process uses the cell location annotation to remove the distortions by Thin-Plate Spline (TPS)~\cite{thinPlateSpline}~transformation and remove the rotations and inclinations of the table by Affine transformation~\cite{AffineWolfram}. Figure~\ref{fig: table image rectification process} illustrates the main steps of the algorithm. Firstly, we generate a binary map of each cell based on the spatial annotations of a table and apply the morphological closure operation, a classical image processing algorithm, to fill the gap between each cell to obtain a table-region based binary map. Secondly, find the table outline by tracing the border of the binary map via the \textit{findContour} API in OpenCV. Then we find the corner points of the outline (\red{red points} in Figure~\ref{fig: table image rectification process} (e)) and link the corner points to obtain the corner lines (\red{red dashed lines}). The TPS control points (\shallowBlue{blue points}) are the equal-division points of the corner lines. Make lines pass through target points and are perpendicular to the corner lines, and we obtain the normal lines (\blue{blue lines}) whose intersections with the table outline are exactly the source points (\orange{orange points}). The next step uses the TPS transformation to minimize the distances between target points and source points with the smallest bending energy and establishes a coordinate map between the original table image and the transformed table image of which the curvature distortion is eliminated. With the four corner points, the algorithm computes the Affine parameters to transform the image, which can remove the rotation and inclination of the table, and finally, we obtain the rectified table image.

\begin{figure}[htb]
    \begin{center}
       \includegraphics[width=1\linewidth]{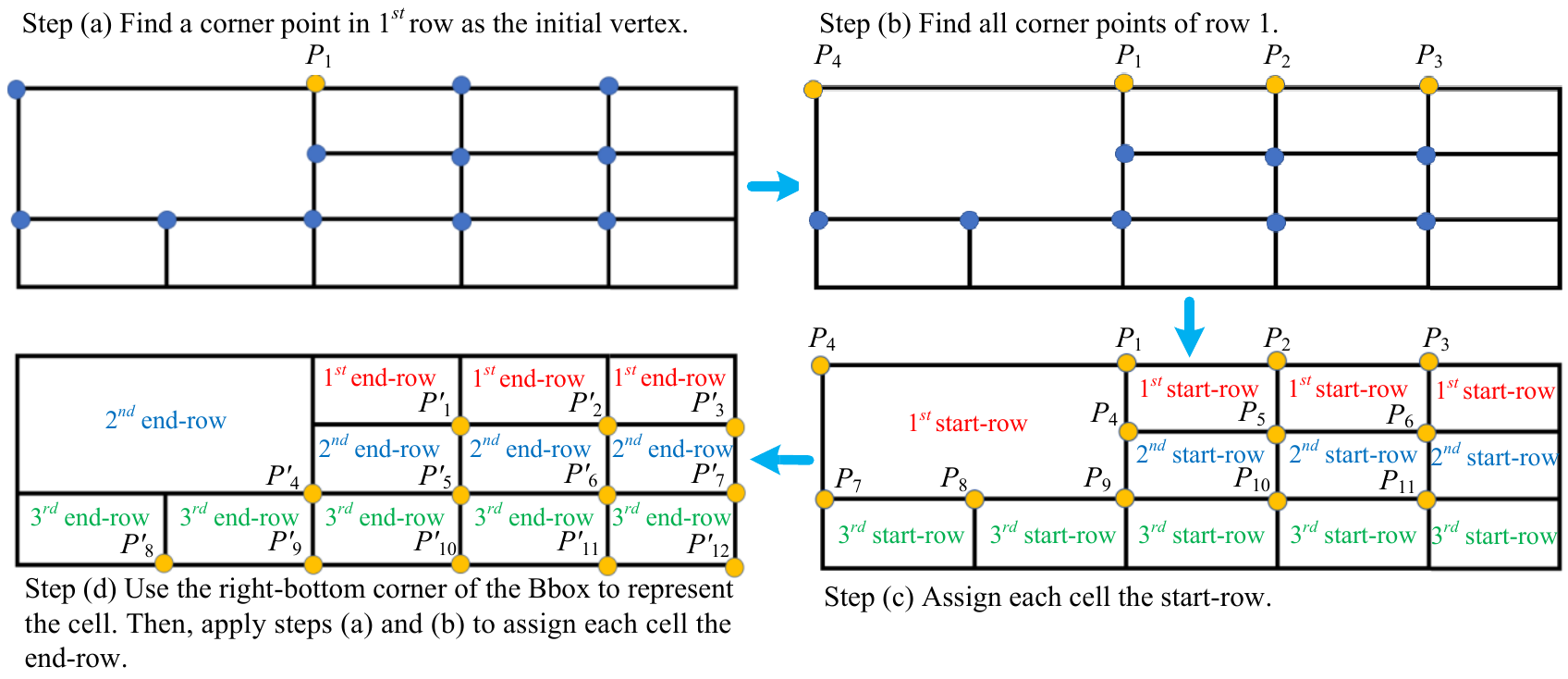}
        \caption{The key steps of our logical property computing algorithm on the rectified table image. We use a regular table on a plain white background to represent the rectified table image.}
        \label{fig: cell logical information computing process}
    \end{center}
\end{figure}

The logical property computing process computes the logical properties by analysing the spatial relationships among cells on the rectified table and assigns the logical properties to the cell on the original table according to the coordinate map from the rectification process. Figure~\ref{fig: cell logical information computing process} shows key steps to analyse the spatial relationship and compute the logical properties:

Step (a), we choose the left-top corner point of a Bbox to represent a cell. The algorithm first finds an initial point $P_1$ of which the $y$-coordinate is the smallest.

Step (b), find all the points satisfying the restriction $|P^{y}_{1}-P^{y}_{2}|<min\{H_{Bbox}\}\times0.4$ which ensures they are in the same row. The $P^{y}$ is the $y$-coordinate, $min\{H_{Bbox}\}$ is the height of the shortest Bbox in the table, and the parameter 0.4 is an empirical constant. Then, we can obtain all cells with the $1^{st}$ start-row (or the $1^{st}$ row in short) and remove them from the table for the next step.

Step (c), repeat Step (a)-(b) until there is no point left on the table, and all cells are assigned with the start-row property.

Step (d), to obtain the end-row property, we can use the right-bottom corner point to represent each cell and do steps (a)-(c).

Step (e), the rowspan of each cell is computed via the formula: rowspan $=$ end-row $-$ start-row $+$ 1.

Step (f), symmetrically, the colspan of each cell is computed via the formula: colspan ($=$ end-column $-$ start-column $+$ 1) after obtaining the start- and end-column of the cells by following (a)-(d) steps on the x-coordinate.

\subsubsection*{Annotation Step}
\label{subsubsec: data annotation step}
According to the data annotation step shown in Figure~\ref{fig: Data creation flow chart} and the intuitive illustration shown in Figure~\ref{fig: Data annotating diagram}, we first draw polygons along the cell borders in the image panel to annotate each cell the spatial location and assign \textit{"group\_id"}s to these polygons for distinguishing table instances using the properties setting panel.

For the efficiency consideration, we apply the TSR auto-annotating algorithm (see the \hyperref[subsubsec: TSR Annotation Generating Algorithm]{TSR Annotation Generating Algorithm} subsection) to generate the logical properties for each cell automatically, then detect the occasional generating errors in the annotation visualization region and manually revise the errors via the properties setting panel. This algorithm has approximately 80\% accuracy, so we only needed to fix the remained 20\% of the annotations, which significantly improved our annotation efficiency.

As shown in Figure~\ref{fig: Data annotating diagram}, for text content transcription, we can directly type in the text for each cell in the digital table shown in the annotation visualization region, and the tool will store the text in the \textit{<Text content>} part of the encoded string in the \textit{"label"} field. Note that for an indistinguishable blurred character, we replace its actual annotation with the \textit{\#} symbol to indicate its existence. For a cell with multiple text lines, we use the \textit{$\backslash$n} escape symbol to separate each line and concatenate the text lines to a single text string.

Finally, we auto-generate the TD annotation for the whole table based on the spatial cell annotations via an image processing program. The program process is as follows: (1) convert spatial annotations (i.e. polygon list) to binary maps, then concatenate them together to obtain a single segmentation map~\cite{DBLP:journals/corr/abs-2001-05566}; (2) use the morphological closure operation on the segmentation map to fill the gaps between each cell's binary region; (3) convert the segmentation map to the polygon along the map contour via the \textit{findContour} API in OpenCV. After the generation, we refine the generated annotation manually in the image panel to ensure the rightness of the annotation.

\subsection*{Border-incomplete Table Generation}
\label{subsec: Data Generation}
As shown in Figure~\ref{fig: Data creation flow chart}, this border-incomplete table generation step aims to produce the three-line and no-line table by erasing the target rule lines of the annotated all-line table.

\begin{figure}[htb]
    \begin{center}
       \includegraphics[width=1\linewidth]{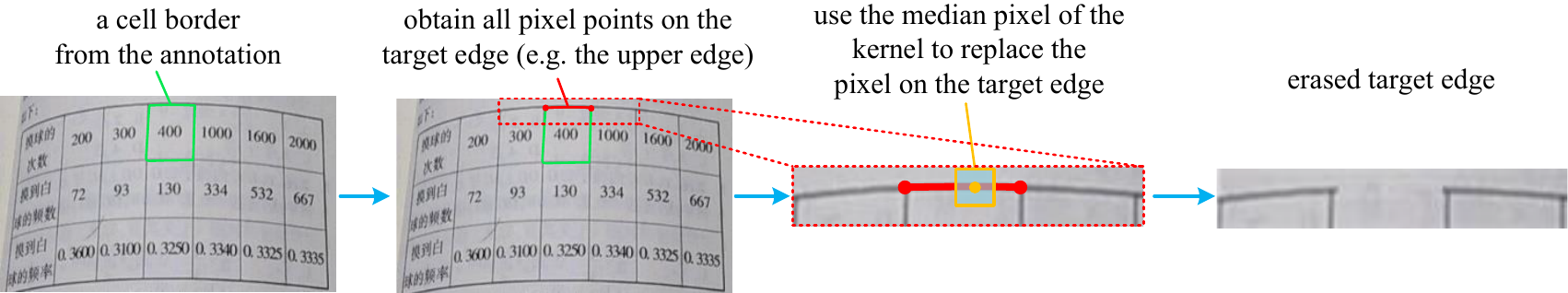}
        \caption{The key steps of erasing a target cell edge. \red{Red points}: Two corner points of the cell border. \red{Red line}: A target edge of a cell. {\color{orange}Orange square}: A kernel centered on a pixel (the {\color{orange}orange point}) on the target edge.}
        \label{fig: target_rule_line_erasing_process}
    \end{center}
\end{figure}

\begin{figure}[htb]
    \begin{center}
       \includegraphics[width=1\linewidth]{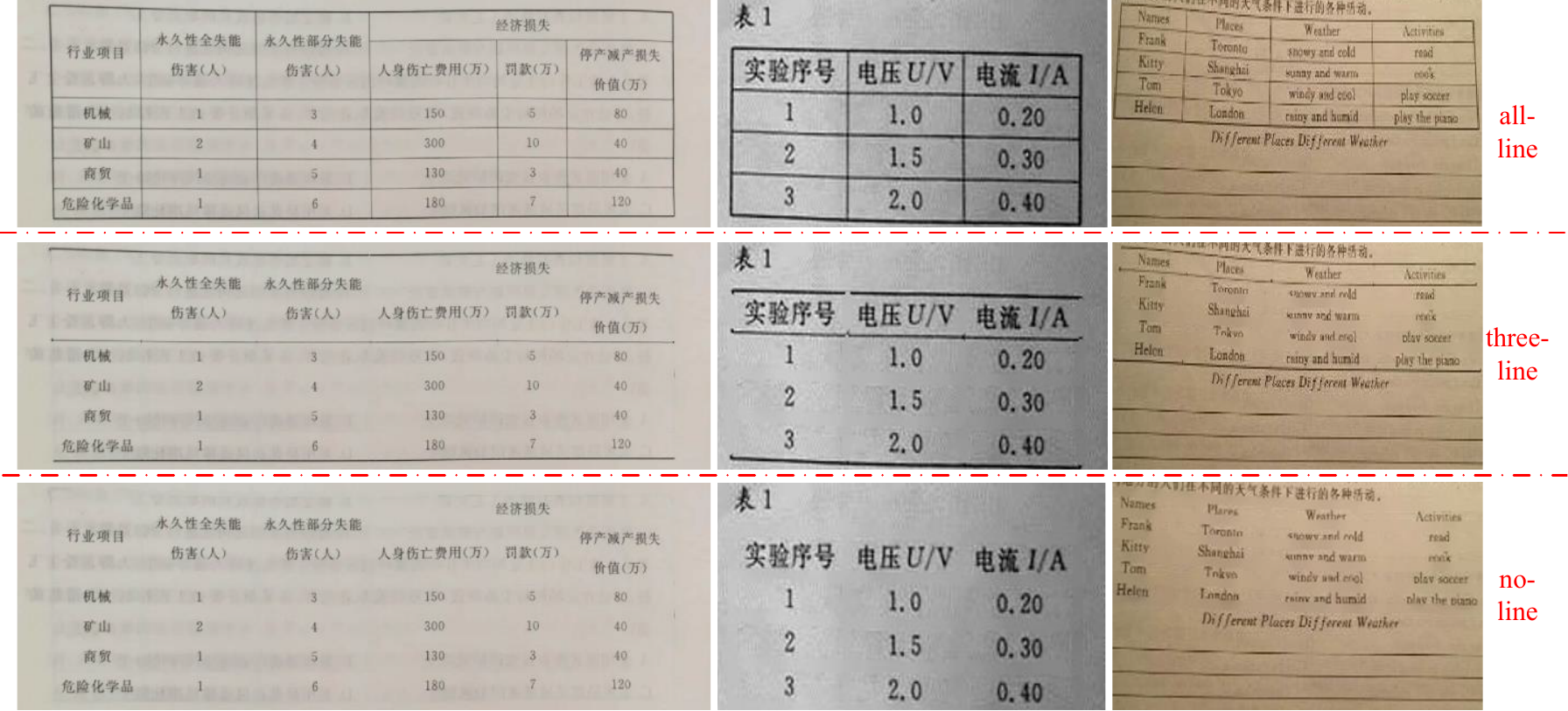}
        \caption{Generated three-line table examples.}
        \label{fig: wire-incomplete table generating result}
    \end{center}
\end{figure}

A table rule line is composed of the edges of the cells on the same row or column, and thus we can erase a table rule line by erasing the cell edges. Since the cell-wise polygons along the cell borders are annotated, we apply an image processing program using these polygon annotations to find the cell edges in the image and remove these edges. Figure~\ref{fig: target_rule_line_erasing_process} shows how the program process erases a single cell edge. Given a table image and the cell border from the annotation, the first step of erasing a cell edge is to obtain all pixel points on the edge (\red{red line}) by extracting the sorted points on the cell border bounded by two adjacent corner points (\red{red points}). For example, the upper edge is a sorted point list starting with the upper-left corner point and ending with the upper-right corner point. The second step is to replace each pixel on the edge with the median pixel of a kernel ({\color{orange}orange square}), which is centered on the pixel ({\color{orange}orange point}). Because the median pixel of the kernel mainly refers to the background color, this step actually replaces the border pixel with the background color. When all pixels on the edge are replaced, this edge can be regarded as removed from the image.

To generate a three-line table, we should erase the horizontal (row) rule lines, ranging from the third to the last but one, and all vertical (column) rule lines. Concretely, we erase the left and right edges of the cells on the $1^{st}$ row and erase all cell edges on other rows except the upper edge on the $2^{nd}$ row and the bottom edge on the last row. As for the no-line table generation, we should erase all target rule lines, and thus we simply erase all edges of every cell in the table. Figure~\ref{fig: wire-incomplete table generating result} illustrates the generating performance. 

Note that the border-incomplete table shares the same cell location annotation with the original all-line table. The polygon annotation is originally drawn along the cell border in the all-line table, while these borders may be erased in the border-incomplete table, so the spatial annotation for a cell is a loose polygon relative to the text content in the border-incomplete table case. Though existing datasets~\cite{DBLP:conf/eccv/ZhongSJ20, DBLP:journals/corr/abs-1908-04729} use a compact Bbox for the text content as the cell location annotation, we insist on our loose annotation because of the existence of the cell borders even though they are invisible (erased). The insight is that a human can somehow infer where are the invisible borders in the image by visual cues or semantic meanings and this insight means the existence and uniqueness of invisible borders in the border-incomplete table. We believe that providing polygon annotation for the invisible borders can facilitate the emergence and development of the cell invisible border recovery task.

\subsection*{Summary of Tools}
\label{subsec: Summary of Tools}

Tab.~\ref{tab: Summary of Tools} is a complete summary of the tools we used during the dataset creation procedure. In the table, we describe the primary uses of these tools and their advantages compared to the alternatives. The last column of the table lists the tool versions, which sometimes matter during the creation procedure.
 
\begin{table}[htb]
    \begin{center}
    \setlength{\tabcolsep}{0.8mm}
        \renewcommand\arraystretch{0.9}
        \begin{tabular}{|c|l|l|c|}
        \hline
        \textbf{Name}                                                                                                     & \multicolumn{1}{c|}{\textbf{Description}}                                                                                                                                                                                                                                                                                                                                          & \multicolumn{1}{c|}{\textbf{Advantages}}                                                                                                                                                                                                                                                                                                                                                                                                                                                  & \textbf{Version} \\ \hline
        \textit{TableMe}                                                                                                       & \begin{tabular}[c]{@{}l@{}}It is our proposed tool for table-specific annotating,\\ which supports annotation for multiple task types,\\ such as table detection, table segmentation, table\\ structure, and table content recognition tasks.\end{tabular}                                                                                           & \begin{tabular}[c]{@{}l@{}}1. It uses interactive visualization to execute\\ the annotation process effectively.\\ 2. Compared to alternatives, it supports the\\ table segmentation task.\end{tabular}                                                                                                                                                                                                                                                                                               & 1.0.0            \\ \hline
        \begin{tabular}[c]{@{}c@{}}\textit{Duplicate}\\ \textit{Image}\\ \textit{Finder}~\cite{Duplicate-Image-Finder}\end{tabular} & \begin{tabular}[c]{@{}l@{}}\textit{Duplicate Image Finder} "looks" at your images to\\ find look-alike images in a folder. It can identify\\ similar and duplicate images even if they are\\ edited, rotated or flipped.\end{tabular} & \begin{tabular}[c]{@{}l@{}}1. It can identify rotated at 90°, 180°, 270°, flipped\\ horizontally and/or vertically duplicate images.\\ 2. It can show all the duplicate images in groups\\ and mark the smaller resolution and/or smaller file\\ size (lower quality ones) images to be deleted.\end{tabular} & 4.8.0          \\ \hline
        \begin{tabular}[c]{@{}c@{}}\textit{Image}-\\ \textit{Assistant}~\cite{imageAssistant}\end{tabular}                 & \begin{tabular}[c]{@{}l@{}}\textit{ImageAssistant} is an extension software running\\ in Chrome and its derivative browsers to analyze\\ and extract pictures in web pages and provide\\ multiple filtering methods to assist users in\\ selecting and downloading.\end{tabular}                                                             & \begin{tabular}[c]{@{}l@{}}Different from browser extensions that provide\\ similar functions in the past, this extension\\ combines multiple data extraction methods to\\ ensure that the images that have appeared can\\ be extracted as comprehensively as possible\\ from various complex structure pages.\end{tabular}                                                                                                                                                                           & 1.66.6           \\ \hline
        \end{tabular}
    \end{center}
    \caption{Summary of the tools we used during the dataset creation procedure}
    \label{tab: Summary of Tools}
\end{table}

\section*{Data Records}


\subsection*{Directory Structure of \textit{TabRecSet}}
\begin{figure}[htb!]
    \begin{center}
       \includegraphics[width=1\linewidth]{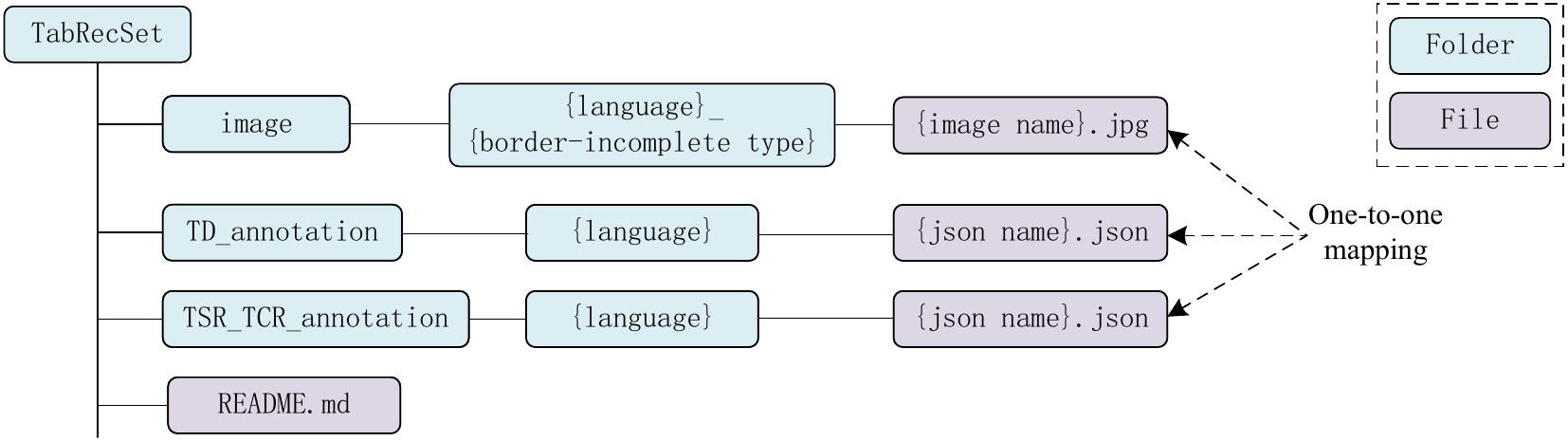}
        \caption{Structure of the data included in \textit{TabRecSet} dataset.}
        \label{fig: directory structure}
    \end{center}
\end{figure}

\textit{TabRecSet} is publicly available in figshare~\cite{TabRecSet}. Its directory structure is shown in Figure~\ref{fig: directory structure}. The image folder contains original and generated table images in JPG format in which tables with different languages and border-incomplete types are separated into corresponding sub-folders. Each image is one-to-one mapped to a unique TD annotation (JSON files in the TD\_annotation folder) and TSR/TCR annotation (JSON files in the TSR\_TCR\_annotation folder) according to its filename, and each JSON file is also divided into different sub-folders by their language. Note that an original all-line table shares the same filename and annotations with its generated border-incomplete tables since the annotations for the all-line table are also valid for the generated images. In the README.md file, we summarise the meta information, such as the dataset license, download links, description of the file format and a link to the source code repository, etc. 

\subsection*{Statistics of \textit{TabRecSet}}
\begin{figure}[htb!]
    \centering
    \includegraphics[width=1\linewidth]{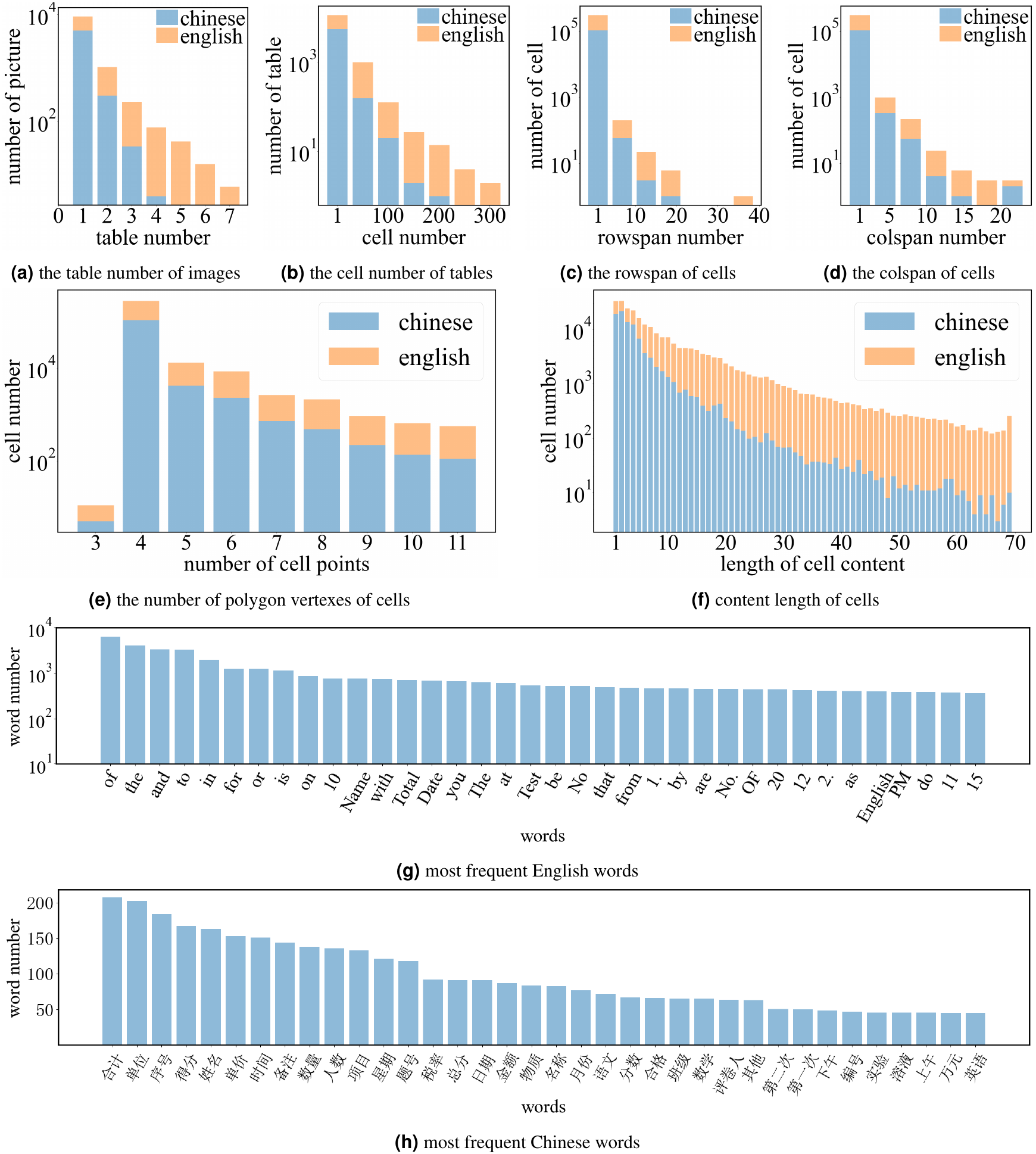}
    \caption{Statistics data of \textit{TabRecSet}.}
    \label{fig: statistics data of TabRecSet}
\end{figure}

To quantitatively verify the data quality and challenge of \textit{TabRecSet} for each sub-task, we analyse its overall size and several instance-wise (e.g., table-wise, cell-wise) statistical characteristics and draw their distributions in Figure~\ref{fig: statistics data of TabRecSet} and Table~\ref{tab: character frequency}. The instance-wise statistical characteristics include the table number of each cell (Figure 11(a)), cell number of each table (Figure 11(b)), rowspan/colspan of each cell (Figure 11(c)/11(d)), vertex number of each cell (Figure 11(e)), content length of each cell (Figure 11(f)),
word frequency (Figure 11(g) and 11(h)) and character frequency (Table~\ref{tab: character frequency}).

In the aspect of overall size, \textit{TabRecSet} contains 32,072 images and 38,177 tables in total among which 16,530 images (17,762 tables) are in Chinese, 15,542 images (20,415 tables) are in English and 21,228 images (25,279 tables) are generated (three-line and no-line). The generated table subset (border-incomplete folder in Figure~\ref{fig: directory structure}) contains 5,113 images and 6728 tables (both three- and no-line tables) in English, 5,501 images and 5,911 tables in Chinese (both three- and no-line tables). 

\begin{table}[htb!]
\setlength{\tabcolsep}{0.8mm}
\renewcommand\arraystretch{1.0}
\begin{center}
\begin{tabular}{|c|c|c|c|c|c|c|c|c|c|c|c|c|c|}
\hline
\textbf{Character}                                                            & A                                                       & B                                                       & C                                                       & D                                                       & E                                                      & F                                                       & G                                                       & H                                                      & I                                                      & J                                                      & K                                                       & L                                                      & M                                                      \\ \hline
\textbf{Count} & \begin{tabular}[c]{@{}c@{}}34,067\\ 1,318\end{tabular}  & \begin{tabular}[c]{@{}c@{}}13,002\\ 989\end{tabular}    & \begin{tabular}[c]{@{}c@{}}25,970\\ 1,499\end{tabular}  & \begin{tabular}[c]{@{}c@{}}18,093\\ 672\end{tabular}    & \begin{tabular}[c]{@{}c@{}}27,634\\ 372\end{tabular}   & \begin{tabular}[c]{@{}c@{}}10,560\\ 423\end{tabular}    & \begin{tabular}[c]{@{}c@{}}9,072\\ 369\end{tabular}     & \begin{tabular}[c]{@{}c@{}}13,034\\ 774\end{tabular}   & \begin{tabular}[c]{@{}c@{}}23,148\\ 296\end{tabular}   & \begin{tabular}[c]{@{}c@{}}3,011\\ 171\end{tabular}    & \begin{tabular}[c]{@{}c@{}}5,444\\ 233\end{tabular}     & \begin{tabular}[c]{@{}c@{}}15,630\\ 598\end{tabular}   & \begin{tabular}[c]{@{}c@{}}19,880\\ 448\end{tabular}   \\ \hline
\textbf{Character}                                                            & N                                                       & O                                                       & P                                                       & Q                                                       & R                                                      & S                                                       & T                                                       & U                                                      & V                                                      & W                                                      & X                                                       & Y                                                      & Z                                                      \\ \hline
\textbf{Count}                                                                & \begin{tabular}[c]{@{}c@{}}22,473\\ 645\end{tabular}    & \begin{tabular}[c]{@{}c@{}}18,390\\ 811\end{tabular}    & \begin{tabular}[c]{@{}c@{}}19,722\\ 492\end{tabular}    & \begin{tabular}[c]{@{}c@{}}1,196\\ 131\end{tabular}     & \begin{tabular}[c]{@{}c@{}}21,210\\ 256\end{tabular}   & \begin{tabular}[c]{@{}c@{}}32,312\\ 616\end{tabular}    & \begin{tabular}[c]{@{}c@{}}25,946\\ 373\end{tabular}    & \begin{tabular}[c]{@{}c@{}}8,149\\ 122\end{tabular}    & \begin{tabular}[c]{@{}c@{}}5,923\\ 304\end{tabular}    & \begin{tabular}[c]{@{}c@{}}6,378\\ 213\end{tabular}    & \begin{tabular}[c]{@{}c@{}}2,249\\ 171\end{tabular}     & \begin{tabular}[c]{@{}c@{}}5,151\\ 163\end{tabular}    & \begin{tabular}[c]{@{}c@{}}951\\ 155\end{tabular}      \\ \hline
\textbf{Character}                                                            & a                                                       & b                                                       & c                                                       & d                                                       & e                                                      & f                                                       & g                                                       & h                                                      & i                                                      & j                                                      & k                                                       & l                                                      & m                                                      \\ \hline
\textbf{Count}                                                                & \begin{tabular}[c]{@{}c@{}}125,450\\ 1,436\end{tabular} & \begin{tabular}[c]{@{}c@{}}25,272\\ 3,285\end{tabular}  & \begin{tabular}[c]{@{}c@{}}50,562\\ 1,105\end{tabular}  & \begin{tabular}[c]{@{}c@{}}49,910\\ 542\end{tabular}    & \begin{tabular}[c]{@{}c@{}}167,771\\ 998\end{tabular}  & \begin{tabular}[c]{@{}c@{}}23,973\\ 233\end{tabular}    & \begin{tabular}[c]{@{}c@{}}33,306\\ 1,826\end{tabular}  & \begin{tabular}[c]{@{}c@{}}42,310\\ 423\end{tabular}   & \begin{tabular}[c]{@{}c@{}}105,066\\ 773\end{tabular}  & \begin{tabular}[c]{@{}c@{}}2,024\\ 29\end{tabular}     & \begin{tabular}[c]{@{}c@{}}12,292\\ 491\end{tabular}    & \begin{tabular}[c]{@{}c@{}}68,205\\ 1,014\end{tabular} & \begin{tabular}[c]{@{}c@{}}43,042\\ 3,023\end{tabular} \\ \hline
\textbf{Character}                                                            & n                                                       & o                                                       & p                                                       & q                                                       & r                                                      & s                                                       & t                                                       & u                                                      & v                                                      & w                                                      & x                                                       & y                                                      & z                                                      \\ \hline
\textbf{Count}                                                                & \begin{tabular}[c]{@{}c@{}}102,827\\ 1,122\end{tabular} & \begin{tabular}[c]{@{}c@{}}111,618\\ 931\end{tabular}   & \begin{tabular}[c]{@{}c@{}}38,488\\ 2,912\end{tabular}  & \begin{tabular}[c]{@{}c@{}}2,101\\ 142\end{tabular}     & \begin{tabular}[c]{@{}c@{}}97,500\\ 577\end{tabular}   & \begin{tabular}[c]{@{}c@{}}100,944\\ 6,096\end{tabular} & \begin{tabular}[c]{@{}c@{}}110,944\\ 1,210\end{tabular} & \begin{tabular}[c]{@{}c@{}}60,738\\ 5,641\end{tabular} & \begin{tabular}[c]{@{}c@{}}13,722\\ 169\end{tabular}   & \begin{tabular}[c]{@{}c@{}}14,070\\ 133\end{tabular}   & \begin{tabular}[c]{@{}c@{}}5,661\\ 868\end{tabular}     & \begin{tabular}[c]{@{}c@{}}24,943\\ 467\end{tabular}   & \begin{tabular}[c]{@{}c@{}}2,255\\ 144\end{tabular}    \\ \hline
\textbf{Character}                                                            & 0                                                       & 1                                                       & 2                                                       & 3                                                       & 4                                                      & 5                                                       & 6                                                       & 7                                                      & 8                                                      & 9                                                      & $\sim$                                                  & `                                                      & !                                                      \\ \hline
\textbf{Count}                                                                & \begin{tabular}[c]{@{}c@{}}72,091\\ 29,532\end{tabular} & \begin{tabular}[c]{@{}c@{}}50,827\\ 22,881\end{tabular} & \begin{tabular}[c]{@{}c@{}}40,859\\ 17,216\end{tabular} & \begin{tabular}[c]{@{}c@{}}23,530\\ 11,122\end{tabular} & \begin{tabular}[c]{@{}c@{}}19,493\\ 9,532\end{tabular} & \begin{tabular}[c]{@{}c@{}}23,627\\ 12,093\end{tabular} & \begin{tabular}[c]{@{}c@{}}15,235\\ 7,452\end{tabular}  & \begin{tabular}[c]{@{}c@{}}13,832\\ 5,595\end{tabular} & \begin{tabular}[c]{@{}c@{}}13,950\\ 6,871\end{tabular} & \begin{tabular}[c]{@{}c@{}}14,121\\ 5,102\end{tabular} & \begin{tabular}[c]{@{}c@{}}65\\ 551\end{tabular}        & \begin{tabular}[c]{@{}c@{}}3,184\\ 1\end{tabular}      & \begin{tabular}[c]{@{}c@{}}253\\ 2\end{tabular}        \\ \hline
\textbf{Character}                                                            & @                                                       & \#                                                      & \$                                                      & \%                                                      & \textasciicircum{}                                     & \&                                                      & *                                                       & (                                                      & )                                                      & -                                                      & \_                                                      & =                                                      & +                                                      \\ \hline
\textbf{Count}                                                                & \begin{tabular}[c]{@{}c@{}}489\\ 13\end{tabular}        & \begin{tabular}[c]{@{}c@{}}34,674\\ 2,289\end{tabular}  & \begin{tabular}[c]{@{}c@{}}1,955\\ 7\end{tabular}       & \begin{tabular}[c]{@{}c@{}}2,058\\ 1,577\end{tabular}   & \begin{tabular}[c]{@{}c@{}}22\\ 5\end{tabular}         & \begin{tabular}[c]{@{}c@{}}1,917\\ 356\end{tabular}     & \begin{tabular}[c]{@{}c@{}}1,341\\ 129\end{tabular}     & \begin{tabular}[c]{@{}c@{}}12,959\\ 3,216\end{tabular} & \begin{tabular}[c]{@{}c@{}}13,420\\ 3,221\end{tabular} & \begin{tabular}[c]{@{}c@{}}21,091\\ 4,141\end{tabular} & \begin{tabular}[c]{@{}c@{}}1,749\\ 439\end{tabular}     & \begin{tabular}[c]{@{}c@{}}1,407\\ 414\end{tabular}    & \begin{tabular}[c]{@{}c@{}}2,426\\ 1,362\end{tabular}  \\ \hline
\textbf{Character}                                                            & \{                                                      & \}                                                      & {[}                                                     & {]}                                                     & |                                                      & \textbackslash{}                                        & /                                                       & \textless{}                                            & \textgreater{}                                         & ,                                                      & .                                                       & ?                                                      & ，                                                      \\ \hline
\textbf{Count}                                                                & \begin{tabular}[c]{@{}c@{}}61\\ 7\end{tabular}          & \begin{tabular}[c]{@{}c@{}}64\\ 7\end{tabular}          & \begin{tabular}[c]{@{}c@{}}461\\ 120\end{tabular}       & \begin{tabular}[c]{@{}c@{}}464\\ 76\end{tabular}        & \begin{tabular}[c]{@{}c@{}}282\\ 24\end{tabular}       & \begin{tabular}[c]{@{}c@{}}241\\ 74\end{tabular}        & \begin{tabular}[c]{@{}c@{}}19,707\\ 7,816\end{tabular}  & \begin{tabular}[c]{@{}c@{}}14,948\\ 5,818\end{tabular} & \begin{tabular}[c]{@{}c@{}}14,886\\ 5,782\end{tabular} & \begin{tabular}[c]{@{}c@{}}16,895\\ 1,076\end{tabular} & \begin{tabular}[c]{@{}c@{}}46,387\\ 11,730\end{tabular} & \begin{tabular}[c]{@{}c@{}}1,485\\ 43\end{tabular}     & \begin{tabular}[c]{@{}c@{}}75\\ 1,598\end{tabular}     \\ \hline
\textbf{Character}                                                            & 。                                                       & ！                                                       & ￥                                                       & （                                                       & ）                                                      & 、                                                       & ：                                                       & ；                                                      & “                                                      & ”                                                      & ？                                                       & 《                                                      & 》                                                      \\ \hline
\textbf{Count}                                                                & \begin{tabular}[c]{@{}c@{}}14\\ 543\end{tabular}        & \begin{tabular}[c]{@{}c@{}}7\\ 20\end{tabular}          & \begin{tabular}[c]{@{}c@{}}1\\ 82\end{tabular}          & \begin{tabular}[c]{@{}c@{}}77\\ 2,657\end{tabular}      & \begin{tabular}[c]{@{}c@{}}104\\ 2,696\end{tabular}    & \begin{tabular}[c]{@{}c@{}}41\\ 1,785\end{tabular}      & \begin{tabular}[c]{@{}c@{}}8,547\\ 2,014\end{tabular}   & \begin{tabular}[c]{@{}c@{}}0\\ 195\end{tabular}        & \begin{tabular}[c]{@{}c@{}}55\\ 128\end{tabular}       & \begin{tabular}[c]{@{}c@{}}20\\ 124\end{tabular}       & \begin{tabular}[c]{@{}c@{}}14\\ 56\end{tabular}         & \begin{tabular}[c]{@{}c@{}}0\\ 132\end{tabular}        & \begin{tabular}[c]{@{}c@{}}1\\ 128\end{tabular}        \\ \hline
\textbf{Character}                                                            & 的                                                       & 一                                                       & 是                                                       & 在                                                       & 不                                                      & 了                                                       & 有                                                       & 和                                                      & 人                                                      & 这                                                      & 中                                                       & 大                                                      & 为                                                      \\ \hline
\textbf{Count}                                                                & \begin{tabular}[c]{@{}c@{}}5\\ 2,818\end{tabular}       & \begin{tabular}[c]{@{}c@{}}11\\ 1,677\end{tabular}      & \begin{tabular}[c]{@{}c@{}}1\\ 294\end{tabular}         & \begin{tabular}[c]{@{}c@{}}3\\ 366\end{tabular}         & \begin{tabular}[c]{@{}c@{}}5\\ 726\end{tabular}        & \begin{tabular}[c]{@{}c@{}}1\\ 152\end{tabular}         & \begin{tabular}[c]{@{}c@{}}3\\ 627\end{tabular}         & \begin{tabular}[c]{@{}c@{}}2\\ 416\end{tabular}        & \begin{tabular}[c]{@{}c@{}}1\\ 1,892\end{tabular}      & \begin{tabular}[c]{@{}c@{}}1\\ 64\end{tabular}         & \begin{tabular}[c]{@{}c@{}}3\\ 924\end{tabular}         & \begin{tabular}[c]{@{}c@{}}1\\ 758\end{tabular}        & \begin{tabular}[c]{@{}c@{}}0\\ 297\end{tabular}        \\ \hline
\textbf{Character}                                                            & 上                                                       & 个                                                       & 国                                                       & 我                                                       & 以                                                      & 要                                                       & 他                                                       & 时                                                      & 来                                                      & 用                                                      & 们                                                       & 生                                                      & 到                                                      \\ \hline
\textbf{Count}                                                                & \begin{tabular}[c]{@{}c@{}}8\\ 830\end{tabular}         & \begin{tabular}[c]{@{}c@{}}4\\ 639\end{tabular}         & \begin{tabular}[c]{@{}c@{}}1\\ 299\end{tabular}         & \begin{tabular}[c]{@{}c@{}}0\\ 170\end{tabular}         & \begin{tabular}[c]{@{}c@{}}9\\ 520\end{tabular}        & \begin{tabular}[c]{@{}c@{}}3\\ 297\end{tabular}         & \begin{tabular}[c]{@{}c@{}}1\\ 161\end{tabular}         & \begin{tabular}[c]{@{}c@{}}7\\ 1,385\end{tabular}      & \begin{tabular}[c]{@{}c@{}}1\\ 107\end{tabular}        & \begin{tabular}[c]{@{}c@{}}1\\ 890\end{tabular}        & \begin{tabular}[c]{@{}c@{}}0\\ 70\end{tabular}          & \begin{tabular}[c]{@{}c@{}}0\\ 935\end{tabular}        & \begin{tabular}[c]{@{}c@{}}3\\ 291\end{tabular}        \\ \hline
\textbf{Character}                                                            & 作                                                       & 地                                                       & 于                                                       & 出                                                       & 就                                                      & 分                                                       & 对                                                       & 成                                                      & 会                                                      & 可                                                      & 主                                                       & 发                                                      & 年                                                      \\ \hline
\textbf{Count}                                                                & \begin{tabular}[c]{@{}c@{}}0\\ 467\end{tabular}         & \begin{tabular}[c]{@{}c@{}}2\\ 665\end{tabular}         & \begin{tabular}[c]{@{}c@{}}0\\ 235\end{tabular}         & \begin{tabular}[c]{@{}c@{}}2\\ 565\end{tabular}         & \begin{tabular}[c]{@{}c@{}}2\\ 63\end{tabular}         & \begin{tabular}[c]{@{}c@{}}0\\ 1,863\end{tabular}       & \begin{tabular}[c]{@{}c@{}}0\\ 285\end{tabular}         & \begin{tabular}[c]{@{}c@{}}2\\ 619\end{tabular}        & \begin{tabular}[c]{@{}c@{}}1\\ 256\end{tabular}        & \begin{tabular}[c]{@{}c@{}}5\\ 254\end{tabular}        & \begin{tabular}[c]{@{}c@{}}2\\ 288\end{tabular}         & \begin{tabular}[c]{@{}c@{}}0\\ 378\end{tabular}        & \begin{tabular}[c]{@{}c@{}}4\\ 1,147\end{tabular}      \\ \hline
\end{tabular}
\end{center}
\caption{Occurring frequency of the most commonly used characters. Each counting data has two values: the upper one is the frequency of the English subset, and the lower one is the Chinese subset.}
\label{tab: character frequency}
\end{table}

We count the table number for every image, which is an image-wise indicator to measure the difficulty of the TD task, and draw the distribution over images in Figure 11(a). According to the statistical result, approximately 300 images in the English subset contain multiple tables, and 1,000 images in the Chinese subset contain multiple tables. The maximum number of tables in an image on English and Chinese subsets are 7 and 4, respectively. We believe that our dataset can benchmark the performance of the TD model in the case that the image contains many tables.

The difficulty of TSR for a table varies with the table size and structure complexity. We choose the cell number as the table-wise indicator and the number of spanning cells as the cell-wise indicator to reflect the two respects, respectively. Figure 11(b) shows the cell number distribution of which the average number is 29 for the English subset, 18 for the Chinese subset, and the maximum number is 351 for the English subset, 207 for the Chinese subset. The spanning cell refers to the cell of which the rowspan or colspan is larger than one, so we summarize the rowspan/colspan of each cell in Figure~11(c)/11(d) to indicate the structure complexity of the dataset. The spanning cell number of the English subset is more than 6,600, and the Chinese subset is more than 2,200. The maximum rowspan and colspan are 41 and 24 for the English subset; 22 and 24 for the Chinese subset. These statistics show that our dataset contains a great number of large tables and has a high overall structure complexity. TSR not only needs to recognize the logical relation among cells but also needs to locate the cell position in the image. The vertex number of a cell polygon reflects the curvature of the cell, which affects how difficult to segment the cell, so we count the distribution of the vertex number for each polygon annotation to verify the challenge of our dataset in cell locating. As shown in Figure~11(e), thousands of cell polygons have more than five vertexes, and hundreds of polygons have more than nine vertexes, which is an extremely distorted case, indicating a high challenge for our dataset in the cell locating task.

Table~\ref{tab: character frequency} exhibits the occurring frequency of commonly used characters in the English and Chinese subsets, which shows the coverage of characters. The commonly used characters include the upper and lower case English letter, digit, English and Chinese punctuation mark, and the most commonly used thirty-two Chinese characters~\cite{commonChinese} (last three rows). Note that we give the frequency of the Chinese character for the English subset and that of the English letter for the Chinese subset because we differ the English and Chinese tables not by the language of the table content but the context, so an English table may contain Chinese characters and vice-versa. Besides the character-wise data for the table content annotation, we also summarize the first thirty-five most occurred words in two subsets, as shown in Figure~11(g)/11(h). Unsurprisingly, the most frequently occurring word in the English subset is mainly prepositions, while the one in the Chinese subset mainly depends on the domain. Table~\ref{tab: character frequency} and Figure~11(g)/11(h) manifest the completeness of the table content annotation, while Figure~11(f) illustrates the content length of each cell, which manifests the difficulty of our dataset in the TCR task.

\section*{Technical Validation}
\label{sec: Technical Validation}

\paragraph{Cross-check:} A total of five qualified persons (including part of the authors) were involved in the dataset creation procedure. We were responsible for data collection, cleaning, annotating, and cross-checking the annotations. As shown in Fig.~\ref{fig: Data annotating diagram}, there are four annotating steps in total. Whenever an annotator finishes an annotating step and is about to move to the next step, he will first exchange his assigned sub-dataset with another annotator and cross-check the annotations.

\paragraph{Proofreading:} After \textit{TabRecSet} was created, a qualified checker (one of the annotators) was designated to filter out or revise bad samples missed to be dealt with during the creation procedure via \textit{TableMe}. There were two types of bad samples to be checked, dirty images (watermarks, sensitive information, etc.) and incorrect annotation, and the checker cancelled the \textit{Check Box} in \textit{TableMe} for the dirty image and revised the incorrect annotation by the tool. After checking a round, the checker filtered out the images for which the \textit{Check Box}es were not checked by a program. Note that we regard the wrongly generated border-incomplete tables, for example, the not fully erased no-line table, as a type of dirty image, so we directly deleted these border-incomplete tables instead of fixing them. Through this round of checking and programming-based filtering procedure, the two types of bad samples were finally cleaned. 

\paragraph{Usability Validation:} To validate the usability of \textit{TabRecSet}, we train or fine-tune a few state-of-the-art methods on our training set (80\% of the whole \textit{TabRecSet}) and evaluate them on the test set (20\%) and record the evaluation results in Tab.~\ref{tab: sota exp}. There is no end-to-end TR model yet, so we validate the usability as completely as possible by covering all sub-tasks.

\begin{table}[htb!]
    \begin{center}
    \setlength{\tabcolsep}{0.8mm}
    \renewcommand\arraystretch{1.2}
    \begin{tabular}{|c|c|c|cc|cc|cc|c|}
    \hline
    \textbf{Model}               & \textbf{\begin{tabular}[c]{@{}c@{}}Support\\ Tasks\end{tabular}} & \textbf{\begin{tabular}[c]{@{}c@{}}Pre-traininig\\ set\end{tabular}} & \multicolumn{1}{c|}{\textbf{\begin{tabular}[c]{@{}c@{}}Training\\ set\end{tabular}}} & \textbf{\begin{tabular}[c]{@{}c@{}}Testing\\ set\end{tabular}} & \multicolumn{1}{c|}{\textbf{\begin{tabular}[c]{@{}c@{}}TEDS-S\\(\%)\end{tabular}}} & \textbf{\begin{tabular}[c]{@{}c@{}}TEDS-All\\ (\%)\end{tabular}} & \multicolumn{1}{c|}{\textbf{\begin{tabular}[c]{@{}c@{}}TSR(-)\\Acc. (\%)\end{tabular}}} & \textbf{\begin{tabular}[c]{@{}c@{}}P-Cell\\(\%)\end{tabular}} & \textbf{\begin{tabular}[c]{@{}c@{}}AP-Table\\(\%)\end{tabular}} \\ \hline
    \multirow{3}{*}{EDD~\cite{DBLP:conf/eccv/ZhongSJ20}}         & \multirow{3}{*}{TSR(-)+TCR}                                      & \multirow{2}{*}{-}                                                   & \multicolumn{1}{c|}{PubTabNet}                                                       & \multirow{3}{*}{\textit{TabRecSet}}                            & \multicolumn{1}{c|}{72.34}                                                             & 50.93                                                               & \multicolumn{1}{c|}{\multirow{6}{*}{NA}}                                              & \multirow{3}{*}{NA}                                                       & \multirow{7}{*}{NA}                                                         \\ \cline{4-4} \cline{6-7}
                                 &                                                                  &                                                                      & \multicolumn{1}{c|}{\multirow{2}{*}{\textit{TabRecSet}}}                             &                                                                & \multicolumn{1}{c|}{51.75}                                                             & 17.04                                                               & \multicolumn{1}{c|}{}                                                                 &                                                                           &                                                                             \\ \cline{3-3} \cline{6-7}
                                 &                                                                  & PubTabNet                                                            & \multicolumn{1}{c|}{}                                                                &                                                                & \multicolumn{1}{c|}{90.68}                                                             & 70.70                                                               & \multicolumn{1}{c|}{}                                                                 &                                                                           &                                                                             \\ \cline{1-7} \cline{9-9}
    \multirow{3}{*}{TableMaster~\cite{DBLP:journals/corr/abs-2105-01848}} & \multirow{3}{*}{TSR}                                             & \multirow{2}{*}{-}                                                   & \multicolumn{1}{c|}{PubTabNet}                                                       & \multirow{3}{*}{\textit{TabRecSet}}                            & \multicolumn{1}{c|}{55.52}                                                             & \multirow{3}{*}{NA}                                                 & \multicolumn{1}{c|}{}                                                                 & 2.974                                                                     &                                                                             \\ \cline{4-4} \cline{6-6} \cline{9-9}
                                 &                                                                  &                                                                      & \multicolumn{1}{c|}{\multirow{2}{*}{\textit{TabRecSet}}}                             &                                                                & \multicolumn{1}{c|}{16.61}                                                             &                                                                     & \multicolumn{1}{c|}{}                                                                 & 0.3524                                                                    &                                                                             \\ \cline{3-3} \cline{6-6} \cline{9-9}
                                 &                                                                  & PubTabNet                                                            & \multicolumn{1}{c|}{}                                                                &                                                                & \multicolumn{1}{c|}{93.13}                                                             &                                                                     & \multicolumn{1}{c|}{}                                                                 & 11.00                                                                     &                                                                             \\ \cline{1-9}
    TGRNet~\cite{DBLP:journals/corr/abs-2106-10598}                       & TSR                                                              & \multirow{2}{*}{-}                                                   & \multicolumn{1}{c|}{\textit{TabRecSet}}                                              & \textit{TabRecSet}                                             & \multicolumn{2}{c|}{\multirow{2}{*}{NA}}                                                                                                                     & \multicolumn{1}{c|}{65.66}                                                            & 74.82                                                                     &                                                                             \\ \cline{1-2} \cline{4-5} \cline{8-10} 
    CDeC-Net~\cite{DBLP:conf/icpr/AgarwalMJ20}                     & TD                                                               &                                                                      & \multicolumn{2}{c|}{\textit{TabRecSet}}                                                                                                               & \multicolumn{2}{c|}{}                                                                                                                                        & \multicolumn{2}{c|}{NA}                                                                                                                                           & 92.80                                                                       \\ \hline
    \end{tabular}
    \end{center}
    \caption{Evaluation results of state-of-the-art methods on \textit{TabRecSet}. \textbf{TSR(-)}: Table topology structure recognition without detecting the cell spatial locations. \textbf{TEDS}: Tree-Edit-Distance-based Similarity~\cite{DBLP:conf/eccv/ZhongSJ20} (TEDS) metric for the topology structure or content recognition. \textbf{TEDS-S}: The TEDS result for TSR(-). \textbf{TEDS-All}: The TEDS result for both TSR and TCR. \textbf{TSR(-) Acc.}: The classification accuracy of logical properties for TSR(-). \textbf{P-Cell}: The precision~\cite{DBLP:journals/ijcv/LiuOWFCLP20} (P) of cell detection. \textbf{AP-Table}: Average Precision~\cite{DBLP:journals/ijcv/LiuOWFCLP20} (AP) of table segmentation. \textbf{NA}: Not applicable.}
    \label{tab: sota exp}
\end{table}

For the topology structure recognition and content recognition tasks, we choose EDD~\cite{DBLP:conf/eccv/ZhongSJ20}~as the baseline model, which is only the model that supports TCR so far. It predicts tables’ Hyper Text Markup Language (HTML) sequences as the results. This HTML sequence contains the table’s topology structure (without cell location) and text content information and can be obtained by converting our annotation~\cite{DBLP:journals/corr/abs-2106-10598}. We choose Tree-Edit-Distance-based Similarity (TEDS)~\cite{DBLP:conf/eccv/ZhongSJ20}~as the metric, which compares the similarity between two tables’ HTML sequences. This metric supports evaluating topology structure or content recognition performance according to whether the sequence contains the text content. Tab.~\ref{tab: sota exp}~shows that EDD fine-tuned on TabRecSet can achieve significantly higher TEDS scores on structure (72.34\%$\rightarrow$90.68\%) and content recognition (50.93\%$\rightarrow$70.70\%). This result illustrates that our training set can help EDD improve the performances and thus validate the usability of \textit{TabRecSet} for these two sub-tasks. As for direct training on our dataset, the performances of EDD are limited (51.75\%, 17.04\%), which reveals that the TSR and TCR tasks on \textit{TabRecSet} are challenging.

For the TSR task (spatial \& topological structure), we choose TableMaster~\cite{DBLP:journals/corr/abs-2105-01848} and TGRNet~\cite{DBLP:journals/corr/abs-2106-10598} as baselines. Both TableMaster and TGRNet output Bboxes as cell detection results, so we use Bbox precision~\cite{DBLP:journals/ijcv/LiuOWFCLP20}~to evaluate the performance of spatial structure recognition. As for the topology structure, there is a difference between the output of the two models. TableMaster output the tables’ HTML sequence as the topology structure prediction, while the TGRNet model formulates the TSR task as a classification problem for each graph node and outputs the Table Graph as the prediction result. In other words, TableMaster and TGRNet represent two different categories of methods to deal with this task, i.e., sequence-based and graph-based methods. The metrics for these two kinds of models are different. The sequence-based model uses TEDS, and the graph-based model uses classification accuracy of the logical properties. As illustrated in Tab.~\ref{tab: sota exp}, TGRNet can achieve moderately high performance on TSR (74.82\% \& 65.66\%). TableMaster with fine-tuning can achieve much higher TSR performance than without fine-tuning (55.52\%$\rightarrow$93.13\%, 2.974\%$\rightarrow$11.00\%). These experiment results are strong evidence of usability for the TSR sub-task. TableMaster without pre-training has a TEDS score of 16.61\%, which reveals the challenge of our dataset for sequence-based methods on topology structure recognition. Note that the performances of TableMaster for spatial structure recognition are low (2.974\%, 0.3524\% and 11.00\%) because TableMaster belongs to the regression-based method, which cannot precisely predict the cell location when the table has a large distortion.

CDeC-Net~\cite{DBLP:conf/icpr/AgarwalMJ20}~is used to verify the usability of our dataset for the TD task. Table~\ref{tab: sota exp} illustrates that the Average Precision~\cite{DBLP:journals/ijcv/LiuOWFCLP20}~(AP) is high enough (92.8\%) to prove the usability for table detection and segmentation.

\section*{Usage Notes}

The data is organized as shown in Fig. \ref{fig: directory structure}. We provide a Python script to load the samples from \textit{TabRecSet} and organize them in a proper data structure. For deep learning research, it is suggested to combine and mix different types or scenarios of tables at first, according to the task needs, and divide the mixed datasets into training, validation, and testing sets for model training, validating, and testing.

\section*{Code Availability}
A link to the dataset, along with Python codes that are used to create the dataset, statistical analysis and plots, is released and publicly available at \href{https://github.com/MaxKinny/TabRecSet}{https://github.com/MaxKinny/TabRecSet}.

\bibliography{main}


\section*{Acknowledgements} 
The research is partially supported by National Nature Science Foundation of China (No. 62176093, 61673182), Key Realm R \& D Program of Guangzhou (No. 202206030001), Guangdong Basic and Applied Basic Research Foundation (No. 2021A151501h2282).

\section*{Author contributions}
Fan Yang originated the concept of this study, designed the study, wrote the codes, annotated the data, and wrote the manuscript. Lei Hu helped with the study designing, data annotating and coding. Xinwu Liu reviewed and revised the manuscript. Shuangping Huang reviewed and revised the manuscript and supervised the study. Zhenghui Gu reviewed and revised the manuscript.

\section*{Competing interests}
The authors declare no competing interests.

\end{CJK*}
\end{document}